\documentclass[journal]{IEEEtran}

\ifCLASSINFOpdf
  \usepackage[pdftex]{graphicx}
\else
\fi
\usepackage[cmex10]{amsmath}
\usepackage{amssymb}
\usepackage{amsthm}
\usepackage{nicealgo}
\usepackage{color}
\usepackage{multicol}
\usepackage{mathtools}
\newtheorem{theo}{Theorem}
\newtheorem{defi}{Definition}
\newtheorem{prop}{Proposition}

\usepackage{url}
\begin{document}

\title{An Iterative Spanning Forest Framework for Superpixel Segmentation}

\author{John~E.~Vargas-Mu{\~{n}}oz,
        Ananda~S.~Chowdhury,~\IEEEmembership{Senior Member,~IEEE},
	      Eduardo~B.~Alexandre,
	      Felipe L. Galv\~{a}o,
        Paulo~A.~Vechiatto~Miranda,
        and~Alexandre~X.~Falc{\~{a}}o, ~\IEEEmembership{Member,~IEEE}
\thanks{John E. Vargas Mu{\~{n}}oz, Felipe L. Galv\~{a}o and Alexandre X. Falc{\~{a}}o are with the Dept. of Information Systems, Institute of Computing, University of Campinas, Campinas, Brazil.
E-mail: afalcao@ic.unicamp.br (corresponding author)}
\thanks{Ananda~S.~Chowdhury is with the Dept. of Electronics and Telecommunication Engineering, Jadavpur University, Kolkata, India.}
\thanks{Eduardo~B.~Alexandre and Paulo~A~Vechiatto~Miranda are with the Dept. of Computer Science, Institute of Mathematics and Statistics, University of S\~{a}o Paulo, S\~{a}o Paulo, Brazil.}}

\IEEEtitleabstractindextext{%
\begin{abstract}
Superpixel segmentation has become an important research problem in
image processing. In this paper, we propose an Iterative Spanning
Forest (ISF) framework, based on sequences of Image Foresting
Transforms, where one can choose i) a seed sampling strategy, ii) a
connectivity function, iii) an adjacency relation, and iv) a seed
pixel recomputation procedure to generate improved sets of connected
superpixels (supervoxels in 3D) per iteration. The superpixels in ISF
structurally correspond to spanning trees rooted at those seeds. We
present five ISF methods to illustrate different choices of its
components. These methods are compared with approaches from the
state-of-the-art in effectiveness and efficiency. The experiments
involve 2D and 3D datasets with distinct characteristics, and a high
level application, named \emph{sky image segmentation}. The
theoretical properties of ISF are demonstrated in the supplementary
material and the results show that some of its methods are competitive
with or superior to the best baselines in effectiveness and
efficiency.
\end{abstract}

}

\maketitle

\IEEEdisplaynontitleabstractindextext

\IEEEpeerreviewmaketitle

\section{Introduction}

\IEEEPARstart{S}{uperpixels} has emerged as an important topic in
image processing. They group pixels into perceptually meaningful
atomic regions \cite{Achanta2012}. A superpixel can be conceived as a
region of similar and connected pixels. Since all the pixels in the
same superpixel exhibit similar characteristics, superpixel primitives
are computationally much more efficient than their pixel
counterparts. It is also expected that the image objects be defined by
the union of their superpixels. Satisfied this property,
superpixels can be used for a variety of applications: medical image
segmentation \cite{Wu2014}, sky segmentation \cite{SkySegmentation},
motion segmentation \cite{Ayvaci2009}, multi-class object segmentation
\cite{Fulkerson2009}, \cite{Yang2012}, object detection
\cite{Shu2013}, spatiotemporal saliency detection \cite{Liu2014},
target tracking \cite{Yang2014}, and depth estimation
\cite{Zitnick2007}.

\begin{figure*}[!htb]
\begin{center}
\begin{tabular}{ccc} 
  \includegraphics[width=5.5cm]{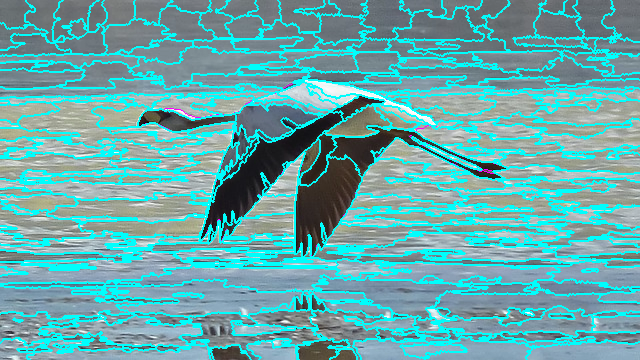} & 
  \includegraphics[width=5.5cm]{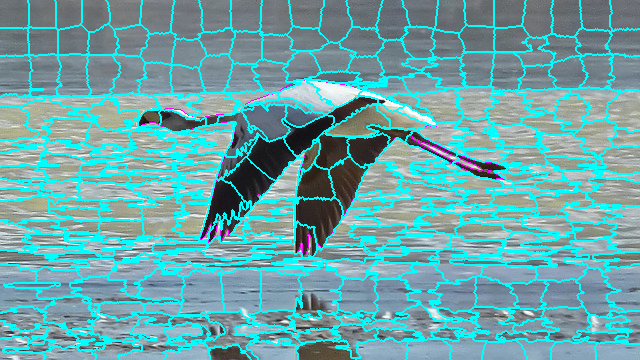} & 
  \includegraphics[width=5.5cm]{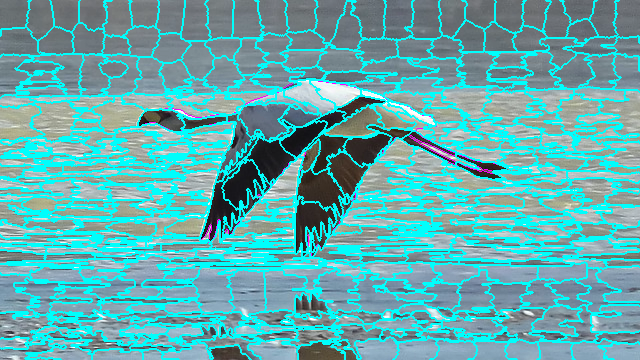} \\ 
(a) & (b) & (c) 
\end{tabular} 
\end{center}
\caption{Superpixel segmentation with the same input number (300) of
  regions for (a) an ISF method, (b) SLIC, and (c) LSC. The borders of the superpixels (cyan) overlap the ground-truth borders (magenta) --- i.e., errors appear in magenta.}
\label{MOTIV}
\end{figure*}

In this paper, we propose an Iterative Spanning Forest (ISF) framework
for generating connected superpixels with no overlap, conforming to a
hard segmentation. Our framework is based on sequences of Image
Foresting Transforms (IFTs)~\cite{Falcao2004} and has four components,
namely, i) a seed sampling strategy, ii) a connectivity function, iii)
an adjacency relation, and iv) a seed recomputation procedure. Each
iteration of the ISF algorithm executes one IFT from a distinct seed
set, yielding to a sequence of segmentation results that improve along
the iterations until convergence. In order to illustrate the
framework, we present i) a mixed seed sampling strategy based on
normalized Shannon entropy, the standard grid sampling, and a
regional-minima-based sampling; ii) three connectivity functions; iii)
two adjacency relations, 4-neighborhood in 2D and 6-neighborhood in
3D; and iv) two seed recomputation procedures. The mixed sampling
strategy aims at estimating higher number of seeds in more
heterogeneous regions in order to improve boundary adherence. Grid
sampling tends to produce more regularly distributed superpixels and
the regional-minima-based strategy aims at solving superpixel
segmentation in a single IFT iteration. Two connectivity functions
allow to control the balance between boundary adherence and superpixel
regularity, and the third one maximizes boundary adherence regardless
to superpixel regularity. Both adjacency relations guarantee the
connectivity between pixels and their corresponding seeds (i.e., a
result consistent with the superpixel definition). For seed
recomputation, we present procedures that exploit color and spatial
information, and color information only. At each iteration, the IFT
algorithm propagates paths from each seed to pixels that are more
closely connected to that seed than to any other, according to a given
connectivity function. The resulting superpixels are spanning trees
rooted at those seeds.

Boundary adherence and superpixel regularity are inversely related
properties. Some works have mentioned the importance of superpixel
regularity --- i.e., of obtaining compact~\cite{Schick2012} and
regularly distributed~\cite{Shi2000,Moore2008} superpixels. However,
the need for superpixel regularity in high level applications requires
a more careful study. Given that the image objects must be
represented by the union of its superpixels, boundary adherence is
certainly the most important property. Figure
\ref{MOTIV} shows segmentation results with the same input number
(300) of superpixels for one of the ISF methods, SLIC (Simple Linear
Iterative Clustering)~\cite{Achanta2012}, and LSC (Linear Spectral
Clustering)~\cite{Chen_2017}. Note that ISF can preserve more
accurately the object borders as compared to SLIC and LSC.

For validation, we first select four 2D image datasets that represent
scenarios with distinct characteristics. The ISF methods are compared
with five approaches from the state-of-the-art:
SLIC~\cite{Achanta2012}, LSC~\cite{Chen_2017}, ERS (Entropy Rate
Superpixel)~\cite{Liu2011}, LRW (Lazy Random Walk)~\cite{Shen2014},
and Waterpixels~\cite{Machairas_2015}. We also add a hybrid approach
that combines ISF with the fastest among them, SLIC. Effectiveness is
evaluated by plots with varying number of superpixels of the most
commonly used boundary adherence measures in superpixel segmentation:
Boundary Recall (BR), as implemented in~\cite{Achanta2012}, and
Undersegmentation Error (UE), as implemented
in~\cite{Neubert2012}. Since SLIC is the only baseline with 3D
implementation, we compare the effectiveness of ISF, SLIC, and the
hybrid SLIC-ISF on the 3D segmentation of three objects --- left brain
hemisphere, right brain hemisphere, and cerebellum --- from volumetric
MR (Magnetic Resonance) images. This experiment uses the most
effective ISF method for this application and effectiveness is
measured by f-score for three segmentation resolutions (low, medium,
and high numbers of supervoxels). Another experiment involves a high
level application, named \emph{sky image segmentation}, in which the
label assignment to the superpixels is decided by an independent and
automatic algorithm. We measure the f-score for varying number of
superpixels using SLIC (the fastest baseline), LSC (the most
competitive baseline), and the most effective ISF method for this
application. For efficiency evaluation, we compare the processing time
for varying number of superpixels among one of the ISF methods,
SLIC-ISF, SLIC, and the two most competitive baselines in
effectiveness, LSC and ERS.

\indent In Section \ref{RW}, we discuss the related works. The ISF
framework and five ISF methods are presented in Section \ref{PM}. In
this section, we also present the general ISF algorithm, discusses
implementation issues, and provide the link for its code. Section
\ref{ER} presents the experimental results and the ISF theoretical
properties are demonstrated in the supplementary material. Section
\ref{CFW} states conclusion and discusses future work.

\section{Related Work}
\label{RW}
Most superpixel segmentation approaches adopt a clustering algorithm
and/or a graph-based algorithm to address the problem in one or
multiple iterations of seed estimation. Several of these methods
cannot guarantee connected superpixels: SLIC (Simple Linear Interative
Clustering)~\cite{Achanta2012}, LSC (Linear Spectral
Clustering)~\cite{Chen_2017}, Vcells (Edge-Weighted Centroidal Voronoi
Tessellations)~\cite{Wang2012}, LRW (Lazy Random
Walks)~\cite{Shen2014}, ERS (Entropy Rate Superpixels)~\cite{Liu2011},
and DBSCAN (Density- based spatial clustering of applications with
noise)~\cite{Shen_2016}. Connected superpixels in these methods are
usually obtained by merging regions, as a post-processing step, which
can reduce the number of desired superpixels.

Some representative graph-based algorithms include Normalized Cuts
\cite{Shi2000}, an approach based on minimum spanning tree
\cite{Felzenszwalb2004}, a method using optimal path via graph cuts
\cite{Moore2008}, an energy minimization framework which can also
yield supervoxels \cite{Veksler2010}, the watershed transform from
seeds \cite{Lotufo2000,Cousty2009,Machairas_2015}, and approaches
based on random walk~\cite{Liu2011,Shen2014}. Normalized cuts can
generate more compact and more regular superpixels. However, as shown
in~\cite{Achanta2012}, it performs below par in boundary adherence
with respect to other methods. The problem with the algorithm in
\cite{Felzenszwalb2004} is exactly the opposite. The resulting
superpixels can conform to object boundaries, but they are very
irregular in size and shape. Similar effect happens in these
graph-based watershed algorithms~\cite{Lotufo2000,Cousty2009}. An
exception is the waterpixel approach~\cite{Machairas_2015} that
enforces compactness by using a modified gradient image. However,
these algorithms try to solve the segmentation problem from a single
seed set (e.g., seeds are selected from the regional minima of a
gradient image). Due to the absence of seed recomputation and/or
quality of the image gradient, they usually miss important object
boundaries. The performance of the method described in
\cite{Moore2008} depends on the pre-computed boundary maps, which is
not guaranteed to be the best in all cases. The authors in
\cite{Veksler2010} actually suggest two methods for generating compact
and constant-intensity superpixels. In~\cite{Liu2011}, the authors use
entropy rate of a random walk on a graph and a balancing term for
superpixel segmentation. The method yields good segmentation results,
but it involves a greedy strategy for optimization. In
\cite{Shen2014}, the authors show that the lazy random walk produces
better results, but the method has initialization and optimization
steps, both requiring the computation of the commute time, which tends
to adversely affect the total execution time.

ISF falls in the category of graph-based algorithms as a particular
case of a more general framework~\cite{Falcao2004} --- the Image
Foresting Transform (IFT). The IFT is a framework for the design of
image operators based on connectivity, such as distance and geodesic
transforms, morphological reconstructions, multiscale skeletonization,
image description, region- and boundary-based image segmentation
methods~\cite{Falcao2000,Lotufo2000,Mansilla2013,Torres2004,Miranda2014,Spina2014,Freitas2016,Falcao2017,Tavares2017},
with extensions to clustering and
classification~\cite{Rocha2009,Papa2009,Papa2012,Amorim2016,
  Papa2017}. As discussed in~\cite{Miranda2009}, by choice of the
connectivity function, the IFT algorithm computes a watershed
transform from a set of seeds that corresponds to a graph cut in which
the minimum gradient value in the cut is
maximized. From~\cite{Cousty2009}, it is known that the watershed
transform from seeds is equivalent to a cut in a minimum-spanning tree
(MST). That is, the removal of the arc with maximum weight from the
single path in the MST that connects each pair of seeds results a
minimum-spanning forest (i.e., a watershed cut). Such a graph cut
tends to be better than the normalized cut in boundary adherence, but
worse in superpixel regularity.

In the evolution of superpixel segmentation methods, it is also worth
mentioning Mean-Shift \cite{ComaniciuM02}, Quick-Shift
\cite{Vedaldi2008}, turbopixels \cite{Levinshtein2009}, SLIC
\cite{Achanta2012}, geometric flow \cite{Wang2013},
LSC~\cite{Chen_2017}, and DBSCAN~\cite{Shen_2016}. The Mean-Shift
method produces irregular and loose superpixels whereas the
Quick-Shift algorithm does not allow an user to choose the number of
superpixels. The turbopixel-based approaches can produce good
superpixels, but are computationally complex. \c{C}i$\check{\rm g}$la
and Atalan \cite{Cigla2010} used connected k-means algorithm with
convexity constraints to achieve superpixel segmentation via
speeded-up turbopixels. The method is still bit slow, and, as claimed
by the authors, fails to provide good boundary recall for complex
images. SLIC is by far the most commonly used superpixel method
\cite{Achanta2012}. It uses a regular grid for seed sampling. Once
chosen, the seeds are transferred to the lowest gradient position
within a small neighborhood. Finally, a modified k-means algorithm is
used to cluster the remaining pixels. This algorithm was shown to
perform better than many other methods (e.g.,~\cite{Levinshtein2009,
  Shi2000, Felzenszwalb2004, Veksler2010, Vedaldi2008}). However, the
k-means algorithm searches for pixels within a $2S \times 2S$ window
around each seed, where $S$ is the grid interval. For a non-regular
seed distribution, some pixels may not be reached by any seed. Indeed,
this might happen from the second iteration on and this labeling
inconsistency problem is only solved by post-processing. In
\cite{Wang2013}, Wang \textit{et al.} proposed a geometric-flow-based
method of superpixel generation. The method has high computational
complexity as it involves computation of the geodesic distance and
several iterations. LSC~\cite{Chen_2017} and DBSCAN~\cite{Shen_2016}
are among the most recent approaches. LSC models the segmentation
problem using Normalized Cuts, but it applies an efficient approximate
solution using a weighted k-means algorithm to generate
superpixels. DBSCAN performs fast pixel grouping based on color
similarity with geometric restrictions, and then merges small clusters
to ensure connected superpixels. 

A first method based on the ISF framework appeared
in~\cite{Alexandre2015} and has been successfully used in a high level
application~\cite{Tavares2017}. It is considered in our experiments.

\section{The ISF framework}
\label{PM}
An ISF method results from the choice of each component: inital seed
selection, connectivity function, adjacency relation, and seed
recomputation strategy. The ISF algorithm is a sequence of Image
Foresting Transforms (IFTs) from improved seed pixel sets (Section
\ref{ift}). For initial seed selection, we propose either grid or
mixed entropy-based seed sampling as effective strategies (Section
\ref{ms}). The closest minima of a gradient image to seeds obtained by
grid sampling is also evaluated in an attempt to solve the problem in
a single iteration. Examples of connectivity functions and adjacency
relations for 2D and 3D segmentations are presented in Sections
\ref{cf} and \ref{ar}, respectively. Two strategies for seed
recomputation are described in Section \ref{sr}. The ISF algorithm is
presented in Section \ref{algo} and its theoretical properties are
demonstrated in the supplementary material. Section~\ref{code}
discusses implementation issues and provides the link to the code.

\subsection{Image Foresting Transform}
\label{ift}
An image can be interpreted as a graph $G=({\cal I},{\cal A})$, whose
pixels in the image domain ${\cal I}\subset Z^n$ are the nodes and
pixel pairs $(s,t)$ that satisfy the \emph{adjacency relation} ${\cal
  A} \subset {\cal I}\times{\cal I}$ are the arcs (e.g., $4$-neighbors
when $n=2$). We use $t\in {\cal A}(s)$ and $(s,t)\in {\cal A}$ to
indicate that $t$ is adjacent to $s$.

\begin{sloppypar}
For a given image graph $G=({\cal I},{\cal A})$, a path $\pi_t=\langle
t_1,t_2,\ldots,t_n = t \rangle$ is a sequence of adjacent pixels with
terminus $t$. A path is \emph{trivial} when $\pi_t=\langle t
\rangle$. A path $\pi_t=\pi_s\cdot \langle s,t\rangle$ indicates the
extension of a path $\pi_s$ by an arc $(s,t)$. When we want to
explicitly indicate the origin of a path, the notation $\pi_{s
  \leadsto t}=\langle t_1 = s,t_2,\ldots,t_n = t \rangle$ is 
used, where $s$ stands for the origin and $t$ for the destination
node.  A \emph{predecessor map} is a function $P$ that assigns to each
pixel $t$ in ${\cal I}$ either some other adjacent pixel in ${\cal
  I}$, or a distinctive marker $nil$ not in ${\cal I}$ --- in which
case $t$ is said to be a \emph{root} of the map. A \emph{spanning
  forest} (image segmentation) is a predecessor map which contains no
cycles --- i.e., one which takes every pixel to $nil$ in a finite
number of iterations. For any pixel $t\in {\cal I}$, a spanning forest
$P$ defines a path $\pi_t^P$ recursively as $\langle t \rangle$ if
$P(t) = nil$, and $\pi_s^P\cdot \langle s,t\rangle$ if $P(t)=s\neq
nil$.
\end{sloppypar}

\begin{sloppypar}
A \emph{connectivity (path-cost) function} computes a value $f(\pi_t)$
for any path $\pi_t$, including trivial paths $\pi_t=\langle t
\rangle$. A path $\pi_t$ is \emph{optimum} if $f(\pi_t) \leq
f(\tau_t)$ for any other path $\tau_t$ in $\Pi_G$ (the set of paths in
$G$).  By assigning to each pixel $t\in {\cal I}$ one optimum path with
terminus $t$, we obtain an optimal mapping $C$, which is uniquely
defined by $C(t) = \min_{\forall \pi_t \;\text{in}\; \Pi_G} \{
f(\pi_t) \}$.  The \emph{Image Foresting Transform}
(IFT)~\cite{Falcao2004} takes an image graph $G=({\cal I},{\cal A})$,
and a connectivity function $f$; and assigns one optimum path $\pi_t$
to every pixel $t \in {\cal I}$ such that an \emph{optimum-path
  forest} $P$ is obtained --- i.e., a spanning forest where all paths
are optimum. However, $f$ must satisfy certain conditions, as
described in~\cite{Ciesielski2016}, otherwise, the paths may not be
optimum.
\end{sloppypar}

In ISF, all seeds are forced to be the roots of the forest by choice
of $f$, in order to obtain a desired number of superpixels. For any
given seed set ${\cal S}$, each superpixel will be represented by its respective
tree in the spanning forest $P$ as computed by the IFT algorithm.

\subsection{Seed Sampling Strategies}
\label{ms}

Any natural image contains a lot of heterogeneity. Some parts of the image can
have really small variations in intensity whereas some parts in the
image can show significant variations. So, it is but natural to choose
more seeds from a more non-uniform region of an image. However, having
a grid structure for the seeds is also essential to conform to the
regularity of the superpixels. The proposed mixed sampling strategy
achieves both the goals. We use a two-level quad-tree representation
of an input 2D image. The heterogeneity of each
quadrant (Q) is captured using Normalized Shannon Entropy
(NSE(Q)). This is given by
\begin{equation}
NSE(Q) = -\frac{\sum_{i=1}^{n}{p_i}{log_2(p_i)}}{log_2n}. 
\end{equation}
Here $n$ denotes the total number of intensity levels in the quadrant
Q and $p_i$ is the probability of occurrence of the intensity $i$ in
the quadrant Q.  For color images, we deem the lightness component in
the Lab color model as the intensity of a pixel. Normalizing the
entropy ensures that the $NSE(Q) \in [0, 1]$. At the first level in
the quad-tree, we compute the normalized Shannon entropies for each
quadrant and also obtain the mean $\mu(NSE)$ and the standard
deviation $\sigma(NSE)$ of the four values. If the value of entropy
for any quadrant exceeds the mean by one standard deviation, i.e., if
$|NSE(Q) - \mu(NSE)| > \sigma(NSE)$, then we further divide the region
in the next level into four quadrants. We then compute the NSE values
for the new quadrants at the second level. Once, the two-level
quad-tree representation is complete, we assign the number of seeds to
be selected from each region as proportional to their NSE values.
Finally, the seeds from each region are picked based on the grid
sampling strategy. So, we essentially perform local grid sampling for
each leaf node in the two-level quad-tree. This procedure may improve
boundary recall with respect to grid sampling, depending on the
dataset. In addition to grid and mixed sampling strategies, we have
also evaluated seed selection based on the reduction of the seed set
generated by grid sampling to the set of the closest regional minima
in a gradient image.

\begin{figure*}[!htb]
\begin{center}
\begin{tabular}{ccccc} 
  \includegraphics[width=3.3cm]{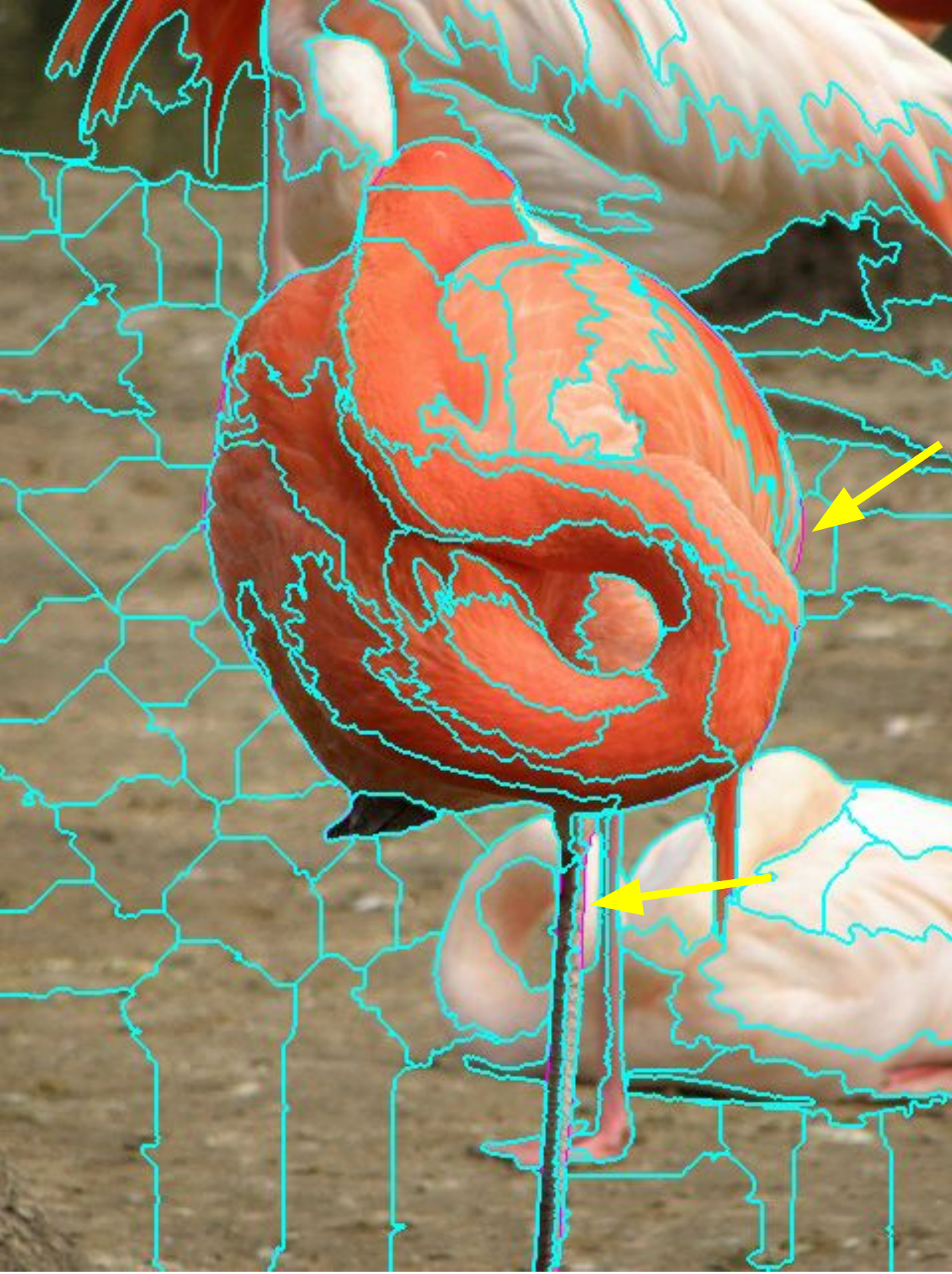}&
  \includegraphics[width=3.3cm]{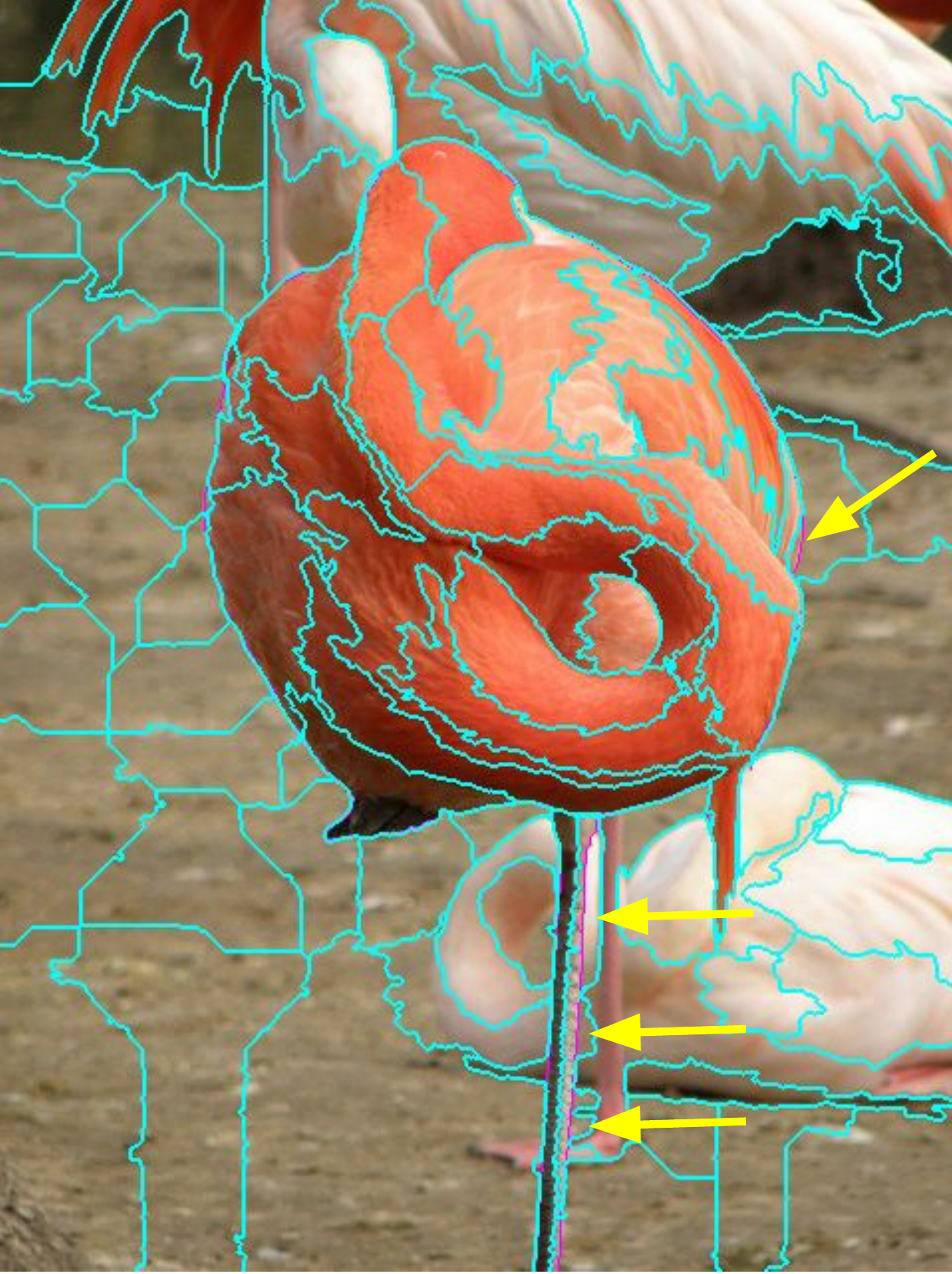}&
  \includegraphics[width=3.3cm]{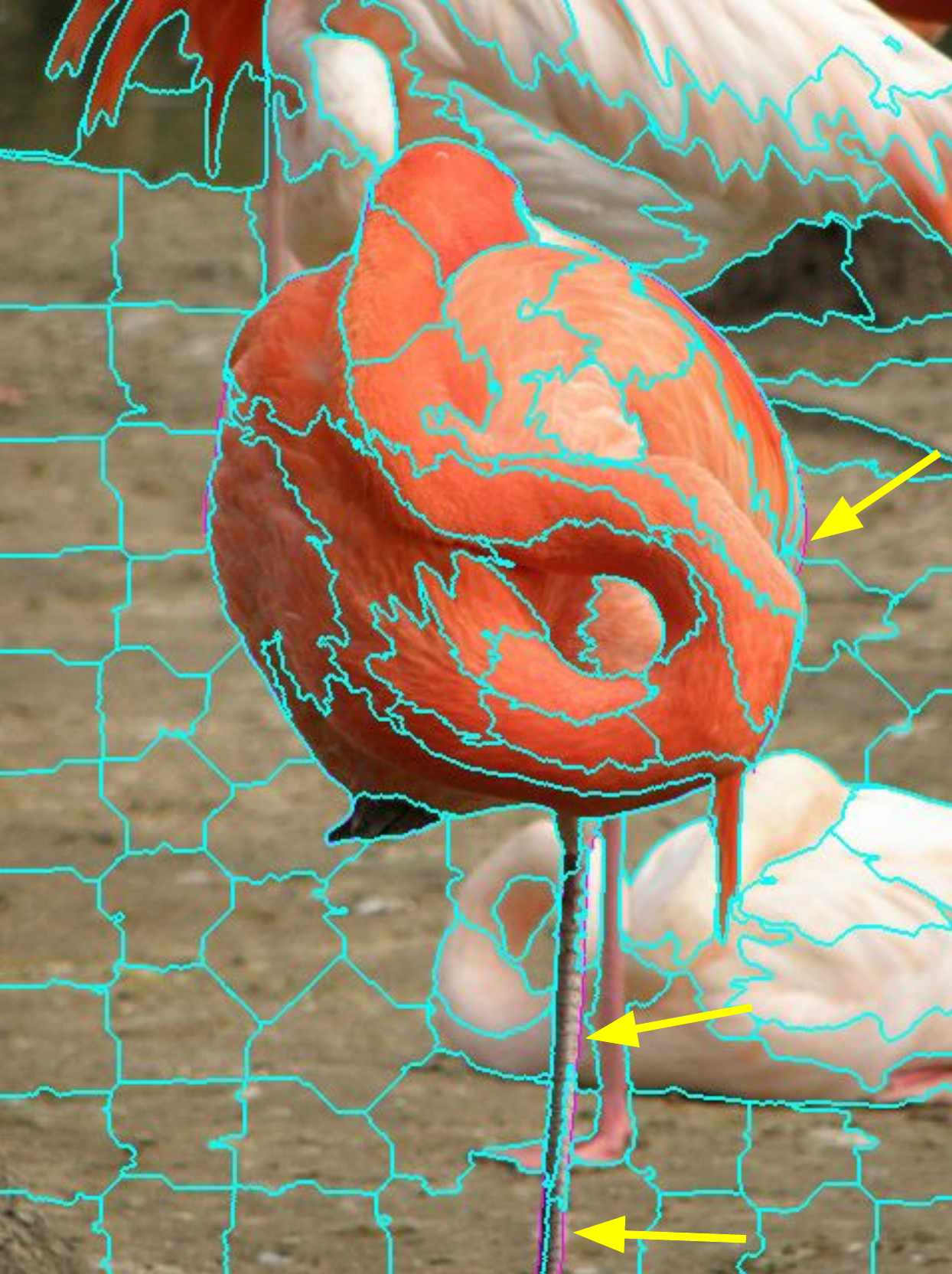}&
  \includegraphics[width=3.3cm]{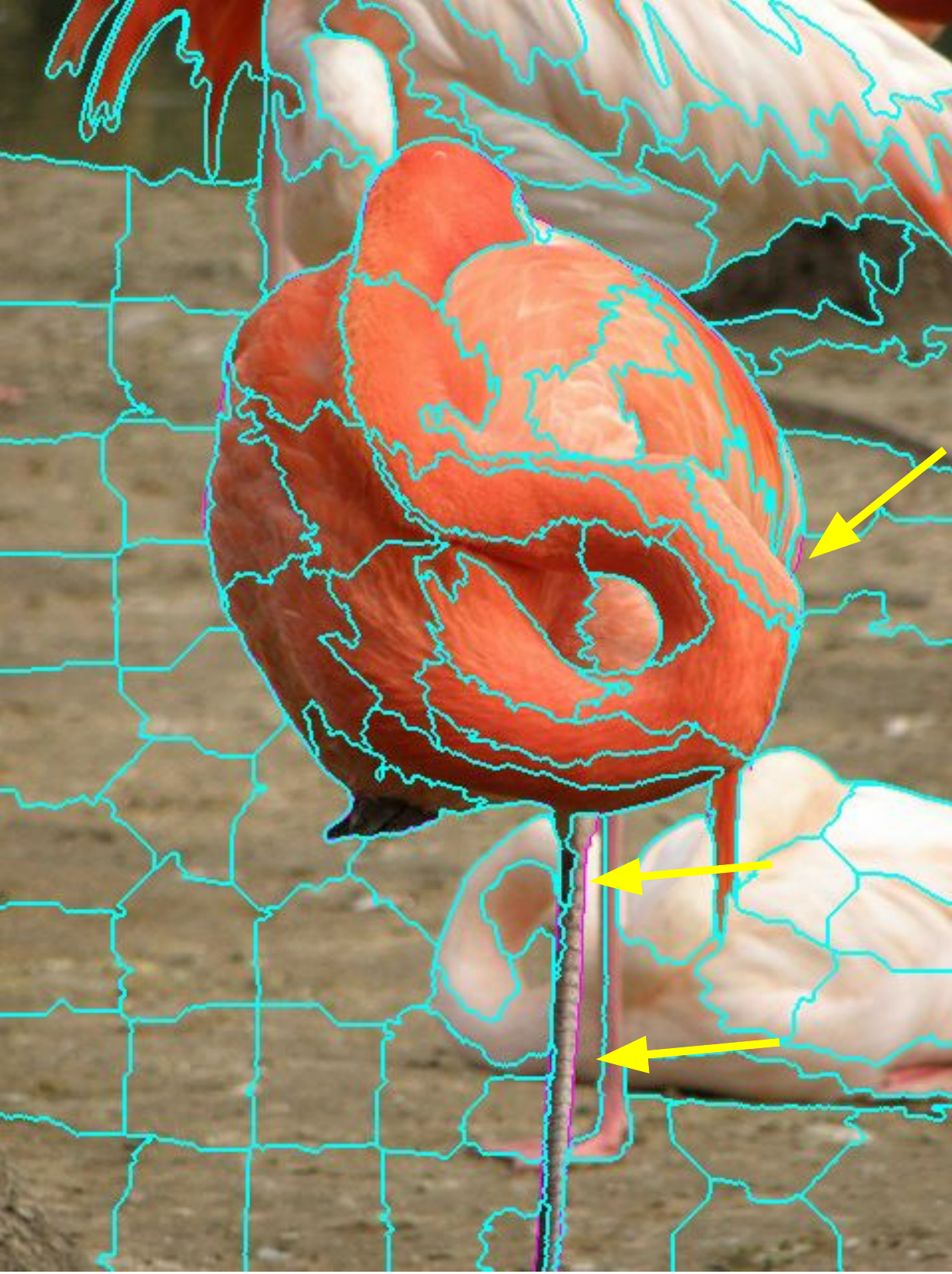}&
  \includegraphics[width=3.3cm]{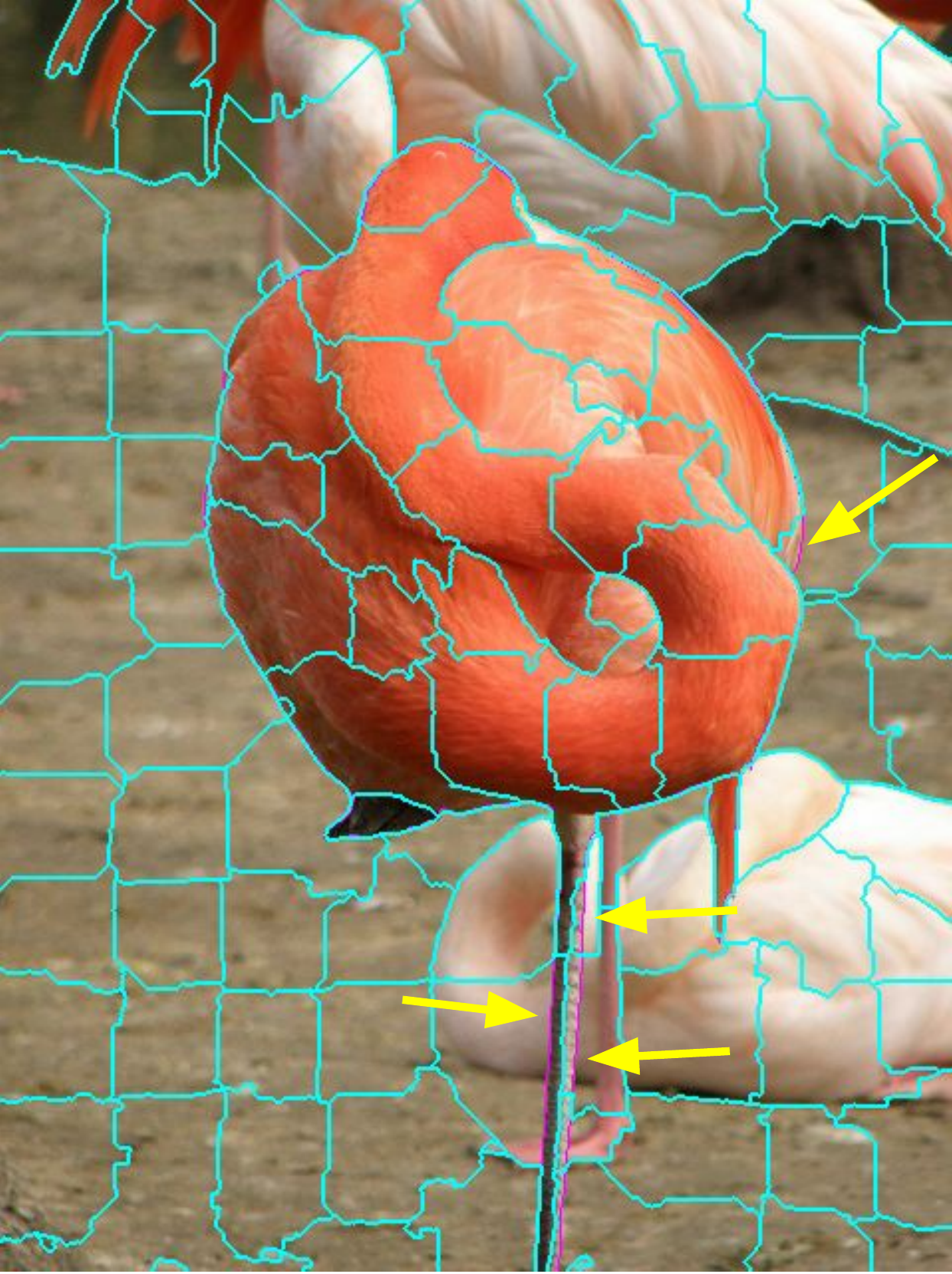}\\
  (a) & (b) & (c) & (d) & (e)
\end{tabular} 
\end{center}
\caption{Segmentation results of an image from
  Birds~\cite{Mansilla2016sib} for five ISF methods (a) ISF-GRID-ROOT
  (BR = 0.93, UE = 0.01), (b) ISF-MIX-ROOT (BR = 0.89, UE = 0.02), (c)
  ISF-GRID-MEAN (BR = 0.90, UE = 0.02), (d) ISF-MIX-MEAN (BR = 0.86,
  UE = 0.02), and (e) ISF-REGMIN (BR = 0.82, UE = 0.02).  Yellow
  arrows indicate leaking between object and background.}
\label{ISF_METHODS_example}
\end{figure*}

\subsection{Connectivity Functions}
\label{cf}
We consider the computation of the IFT with two path-cost functions
that only guarantee a spanning forest, $f_1$ (Equation~\ref{eq_fd1})
and $f_2$ (Equation~\ref{eq_fd2}), and a third one, $f_3$
(Equation~\ref{eq_fd3}), that guarantees an optimum-path forest. The
spanning forest in $f_1$ and $f_2$ might not be optimum, because the
path costs depend on path-root properties~\cite{Ciesielski2016}. However,
these functions can efficiently deal with the problem of intensity
heterogeneity~\cite{Mansilla2013}.

The seed sampling approach (e.g. grid or mixed) defines an initial
seed set ${\cal S}$, such that for each seed pixel $s_j \in {\cal S}$
at coordinate $(x_j, y_j)$, its color representation in the Lab color
space is given $I(s_j)=[l_j\;a_j\;b_j]^T$. A path-cost function $f$ is
defined by a trivial-path cost initialization rule and an
extended-path cost assignment rule. We present three instances of $f$,
denoted as $f_1$, $f_2$ and $f_3$, with trivial-path initialization
rule given by
\begin{eqnarray} 
f_{\ast}(\pi_t = \langle t\rangle) & = & \left\{\begin{array}{ll}
0 & \mbox{if $t\in {\cal S}$,} \\
+\infty & \mbox{otherwise.}
\end{array}\right. 
\label{eq_trivial_path}
\end{eqnarray}

They differ in the extended-path cost assignment rule, as follows. 
\begin{equation}
f_1(\pi_{s_j \leadsto s}\cdot \langle s,t\rangle) = f_1(\pi_s) + \left(\|I(t)-I(s_j) \| \alpha\right)^{\beta} + \|s,t\|,  
\label{eq_fd1}
\end{equation}
where $\alpha \geq 0$, $\beta \geq 1$, and $I(t)=[l_t\;a_t\;b_t]^T$ is
the color vector at pixel $t$. 
\begin{equation}
f_2(\pi_{s_j \leadsto s}\cdot \langle s,t\rangle) = f_2(\pi_s) + \left(\|I(t)-M(s_j) \| \alpha\right)^{\beta} + \|s,t\|,  
\label{eq_fd2}
\end{equation}
where $M(s_j)$ is the mean color, computed inside the superpixel of
the previous iteration, which contains the new seed $s_j$ ($M(s_j) =
I(s_j)$ at the first iteration).
\begin{equation}
f_3(\pi_{r \leadsto s}\cdot \langle s,t\rangle) = \max\{f_3(\pi_s), D(t) \}, 
\label{eq_fd3}
\end{equation}
where $D(t)$ is the value of the gradient image in the pixel $t$.

At the end of the IFT algorithm, each superpixel will be represented
by its respective tree in the spanning forest $P$. After that, an
update step adjusts the roots (new seeds) of the spanning
trees.

For paths $\pi_{t_1 \leadsto t_n}=\langle t_1, t_2, \ldots,
t_n\rangle$, $n > 1$, and additive path-cost function $f(\pi_{t_1
  \leadsto t_n})=\sum_{i=1,2,\ldots,n-1} \{ w(t_i,t_{i+1}) \}$,
$w(t_i,t_{i+1})\geq 0$, the minimization of the cost map imposes too
much shape regularity on superpixels, by avoiding adherence to image
boundaries. On the other hand, $f(\pi_{t_1 \leadsto
  t_n})=\max_{i=1,2,\ldots,n-1} \{ w(t_i,t_{i+1}) \}$
(Equation~\ref{eq_fd3}, for $w(t_i,t_{i+1})=D(t_{i+1})$) provokes high
adherence to image boundaries, but also possible leakings when
delineating poorly defined parts of the boundaries. The path-cost
function $f(\pi_{t_1 \leadsto t_n})=\sum_{i=1,2,\ldots,n-1} \{
w(t_i,t_{i+1})^{\beta} \}$, $\beta > 1$, represents a compromise
between the previous two. We fix $\beta=12$ in all experiments to
approximate the effect of high adherence to image boundaries with
considerably reduced leaking in superpixel segmentation. The arc weight
$w(t_i,t_{i+1})=\| I(t_{i+1}) - I(s_j)\|\alpha$ (Equation~\ref{eq_fd1}
for $s_j=t_1$), or $w(t_i,t_{i+1})=\| I(t_{i+1}) - M(s_j)\|\alpha$
(Equation~\ref{eq_fd2} for $s_j=t_1$), penalizes paths that cross
image boundaries, but the choice of $\alpha$ provides the compromise
between the shape regularity on superpixels, as imposed by the spatial
connectivity component $\|t_{n-1},t_{n}\|$ in Equations~\ref{eq_fd1}
and~\ref{eq_fd2}, and the high boundary adherence of
$\sum_{i=1,2,\ldots,n-1} \{ w(t_i,t_{i+1})^{\beta} \}$ for
$\beta=12$. The choice of $\alpha$ is then optimized to maximize
performance in BR and UE, without compromising too much the regularity
of the superpixels (as it happens with $f_3$).

\subsection{Adjacency Relation}
\label{ar}
The popular choices for adjacency relation are 4- or 8-neighborhood in
2D and 6- or 26-neighborhood in 3D in order to ensure connected
superpixels (supervoxels). We prefer simple symmetric adjacency of
4-neighborhood in 2D and 6-neighborhood in 3D. This choice helps in
the regularity of the superpixels/supervoxels.

\subsection{Seed Recomputation}
\label{sr}
We next discuss the automated seed recomputation strategy.  Let
$s_i^t$ be the $i^{th}$ superpixel root (seed) at iteration $t$ and
its feature vector defined as
$[l_i^t\;a_i^t\;b_i^t\;x_i^t\;y_i^t]^T$. We select $s_i^t$ either as
the pixel of the superpixel whose color is the most similar to the
mean color of the superpixel or as the pixel of the superpixel that is
the closest to its geometric center. During the subsequent IFT
computations, we only recompute the seed $s_i^{t+1}$ if:
\begin{equation}
\| [l_i^t\;a_i^t\;b_i^t]-[l_i^{t+1}\;a_i^{t+1}\;b_i^{t+1}] \| > \sqrt{\mu_c} 
\label{eq_thr_color}
\end{equation}

or
\begin{equation}
\| [x_i^t\;y_i^t]-[x_i^{t+1}\;y_i^{t+1}]  \| > \sqrt{\mu_s},
\label{eq_thr_spatial}
\end{equation}
where $\mu_c$ and $\mu_s$ are the average color and spatial distances
to seed $s_i^{t}$.

\subsection{Five Different ISF Methods}
\label{five_isf_methods}
  
We present five ISF methods. The first two use
function $f_1$, \mbox{ISF-GRID-ROOT} is based on grid sampling and
\mbox{ISF-MIX-ROOT} is based on mixed sampling. They recompute seeds
as the pixel inside each superpixel whose color is the closest to the
mean color of the superpixel. The third and fourth methods use
function $f_2$, \mbox{ISF-GRID-MEAN} is based on grid sampling and
\mbox{ISF-MIX-MEAN} is based on mixed sampling. They recompute seeds
as the pixel inside each superpixel whose position is the closest to
the geometric center of the superpixel.  In~\cite{Alexandre2015}, we
presented \mbox{ISF-GRID-MEAN}.

We now discuss the fifth superpixel generation method, called
\mbox{ISF-REGMIN}, that uses the path-cost function
$f_3$. \mbox{ISF-REGMIN} is designed to be fast, as it uses only a
single iteration of the IFT algorithm with no seed recomputation. This
method initially performs grid sampling to set the seeds. Then, the
seeds are substituted by any pixel at the closest regional minimum,
computed in the gradient image.

It is important to note that the ISF methods do not require a
post-processing step as the connectivity is already guaranteed by
design.

Figure~\ref{ISF_METHODS_example} presents the segmentation results of
the five ISF methods on an image of Birds~\cite{Mansilla2016sib}:
ISF-GRID-ROOT, ISF-MIX-ROOT, ISF-GRID-MEAN, ISF-MIX-MEAN and
ISF-REGMIN. For this dataset, with thin and elongated object parts,
ISF-GRID-ROOT obtains the best result. However, ISF-MIX-MEAN performs
better on most datasets.

\subsection{The ISF Algorithm}
\label{algo}

Algorithm~\ref{alg.isfalgo} presents the Iterative Spanning Forest
procedure. 

\begin{nicealgo}{alg.isfalgo} 
\naTITLE{Iterative Spanning Forest} \naPREAMBLE \naINPUT{Image
  $\hat{I}=({\cal I}, I)$, adjacency relation ${\cal A}$, initial seed
  set ${\cal S}\subset {\cal I}$, the
  parameters $\alpha\geq 0$ and $\beta \geq 1$, and the maximum number
  of iterations $MaxIters \geq 1$.}  \naOUTPUT{Superpixel label map
  $L_s$.}  \naAUX{State map $S$, priority queue $Q$, predecessor map
  $P$, cost map $C$, root map $R$ and superpixel mean color array $M$.
  }  \naBODY \na{$iter \mget 0$}

\naBEGIN{\naWHILE $iter < MaxIter$, \naDO}

\naBEGIN{\naFOREACH $t\in {\cal I}$, \naDO}

\na{$P(t) \mget nil$, $R(t)\mget t$}

\naENDN{1}{$S(t)\mget White$, $C(t)\mget +\infty$}

\na{$label \mget 1$}

\naBEGIN{\naFOREACH $t\in {\cal S}$, \naDO}

\na{$C(t)\mget 0$}

\na{$L_s(t) \mget label$, $label \mget label + 1$}

\na{Insert $t$ in $Q$, $S(t)\mget Gray$}


\naBEGIN{\naIF $iter = 0$, \naTHEN} 

\naENDN{2}{$M(t)\mget I(t)$}

\naBEGIN{\naWHILE $Q\neq \emptyset$, \naDO}

\na{Remove $s$ from $Q$ such that $C(s)$ is minimum}

\na{$S(s) \mget Black$}

\naBEGIN{\naFOREACH $t\in {\cal A}(s)$, such that $S(t) \neq Black$ , \naDO}

\na{$c\mget C(s)+\left(\|I(t)-M(R(s))\|\alpha\right)^{\beta}+\|s,t\|$}

\naBEGIN{\naIF $c < C(t)$, \naTHEN} 

\na{Set $P(t)\mget s$, $R(t)\mget R(s)$}

\na{Set $C(t)\mget c$, $L_s(t) \mget L_s(s)$}

\naBEGIN{\naIF $S(t) = Gray$, \naTHEN}

\naENDN{1}{Update position of $t$ in $Q$}

\naBEGIN{\naELSE}

\na{Insert $t$ in $Q$}

\naENDN{4}{$S(t)\mget Gray$}

\na{${\cal S}, M \mget RecomputeSeeds({\cal S}, \hat{I}, L_s)$}

\naENDN{1}{$iter \mget iter + 1$}

\na{Return $L_s$}

\end{nicealgo}

Line 1 initializes the auxiliary variable $iter$ (iteration number).
The loop in Line 2 stops when the maximum number of iterations is
achieved. Lines 3-5 initialize the values for the predecessor, root,
state and cost maps for all image pixels.  The state map $S$ indicates
by $S(t)=White$ that a pixel $t$ was never visited (never inserted in
the priority queue $Q$), by $S(t)=Gray$ that $t$ has been visited and
is still in $Q$, and by $S(t)=Black$ that $t$ has been processed
(removed from $Q$). Lines 7-12 initialize the cost and label maps and
insert the seeds in $Q$. The seeds are labeled with consecutive
integer numbers in the superpixel label map $L_s$. Lines 13-25 perform
the label propagation process. First, we remove the pixels $s$ that
have minimum path cost in $Q$. Then the loop in Lines 16-25 evaluates
if a path with terminus $s$ extended to its adjacent $t$ is cheaper
than the current path with terminus $t$ and cost $C(t)$.  If that is
the case, $s$ is assigned as the predecessor of $t$ and the root of
$s$ is assigned to the root of $t$ (Line 19). The path cost and the
label of $t$ are updated. If $t$ is in $Q$, its position is updated,
otherwise $t$ is inserted into $Q$.  After the label propagation
stage, function $RecomputeSeeds$ returns the new seed set and the new
mean color values $M$ for the superpixels.  Note that in the first
iteration the feature vector of the superpixel root is the seed pixel
color (Line 11-12).  The tasks of label propagation and seed
recomputation are performed until the condition of Line 2 is
achieved. The algorithm returns the label map $L_s$ (superpixel
segmentation). Note that the algorithm describes the method
ISF-MIX-MEAN if we use mixed sampling as seed initialization strategy.
It uses the path-cost function $f_{2}$ (see Equation~\ref{eq_fd2}) in
Line 17. By replacing Line 17 with the path-cost function $f_{1}$ (see
Equation~\ref{eq_fd1}), we obtain the algorithm for the method
ISF-MIX-ROOT.  Finally, by replacing mixed sampling by grid sampling
in ISF-MIX-ROOT, we obtain the method ISF-GRID-ROOT.

\subsection{Implementation issues and available code}
\label{code}

In general, using a priority queue as a binary heap, each execution of
the IFT algorithm takes time $O(N\log N)$ for $N=|{\cal I}|$ pixels
(linearithmic time). Given that the time to recompute seeds is linear,
the complexity of the ISF framework using a binary heap is
linearithmic, independently of the number of superpixels. For integer
path costs, such as in ISF-REGMIN, it is possible to reduce the IFT
execution time to $O(N)$ using a priority queue based on bucket
sorting~\cite{Falcao2000}.

For efficient implementation, we use a new variant, as proposed in
~\cite{Condori2017}, of the Differential Image Foresting Transform
(DIFT) algorithm~\cite{Falcao2004b}. This algorithm is able to update
the spanning forest by revisiting only pixels of the regions modified
in a given iteration $iter > 1$. The efficient implementation of ISF
is available at
\url{www.ic.unicamp.br/~afalcao/downloads.html}.

\section{Experimental Results}
\label{ER}

In this section, we evaluate the methods based on effectiveness in 2D
and 3D image datasets, effectiveness in a high level application, and
efficiency.

\subsection{Effectiveness in 2D and 3D datasets}
\label{efectiveness}

We first measure the effectiveness of the methods in Boundary Recall
(BR) (as implemented in~\cite{Achanta2012}) and Undersegmentation
Error (UE) (as implemented in~\cite{Neubert2012}) using plots with
varying number of superpixels on four 2D datasets:
Berkeley~\cite{Martin2001} (300 natural images),
Birds~\cite{Mansilla2016sib} (50 natural images),
Grabcut~\cite{Rother_2004} (50 natural images), and Liver (50 CT slice
images of the abdomen). The objects in Birds are fine and elongated
structures and the images of the liver are grayscale.

\begin{figure*}[!htb]
\begin{center}

   \includegraphics[width= 7.7cm]{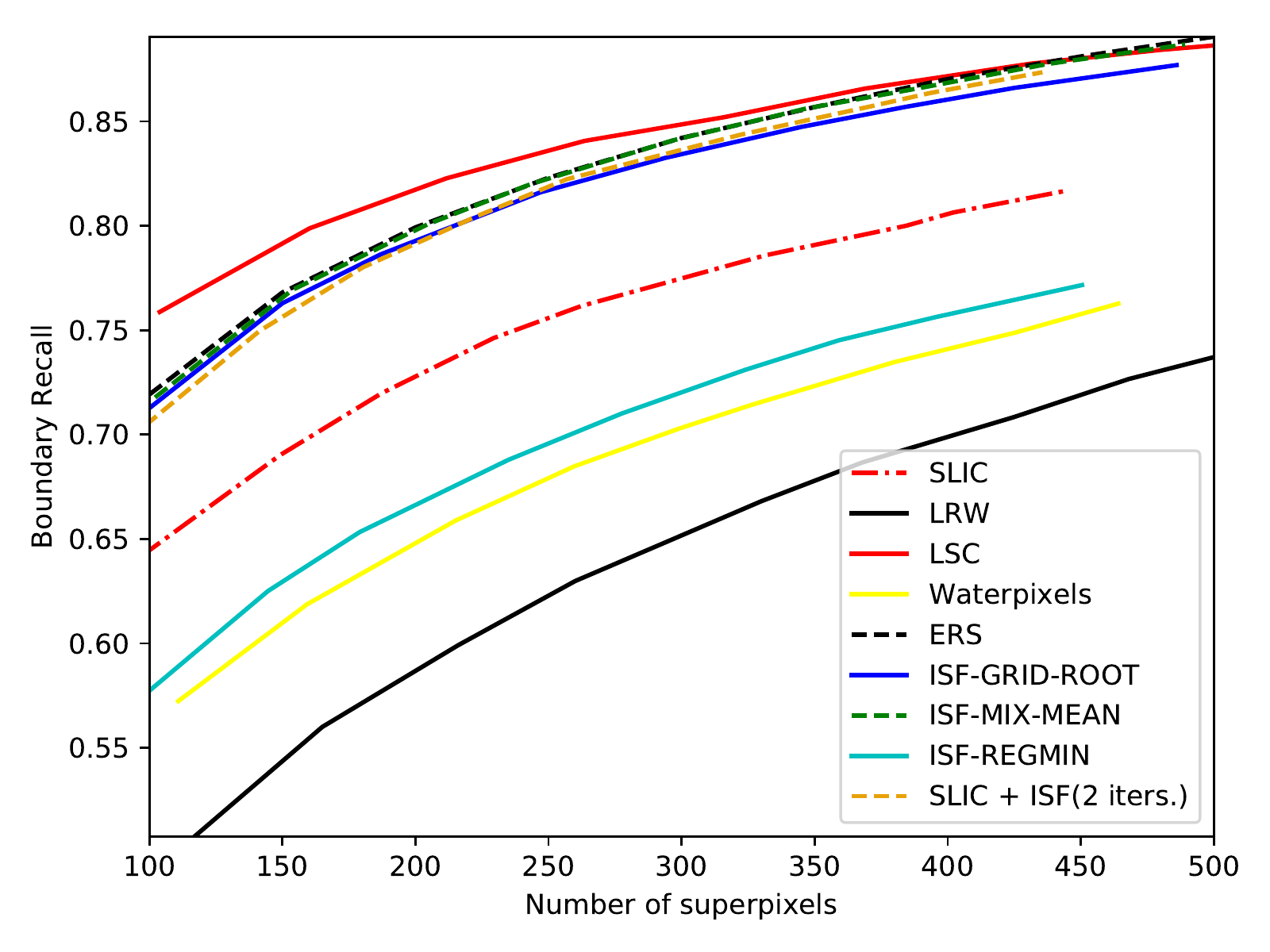} 
	 \hspace{1cm}
	 \includegraphics[width=7.7cm]{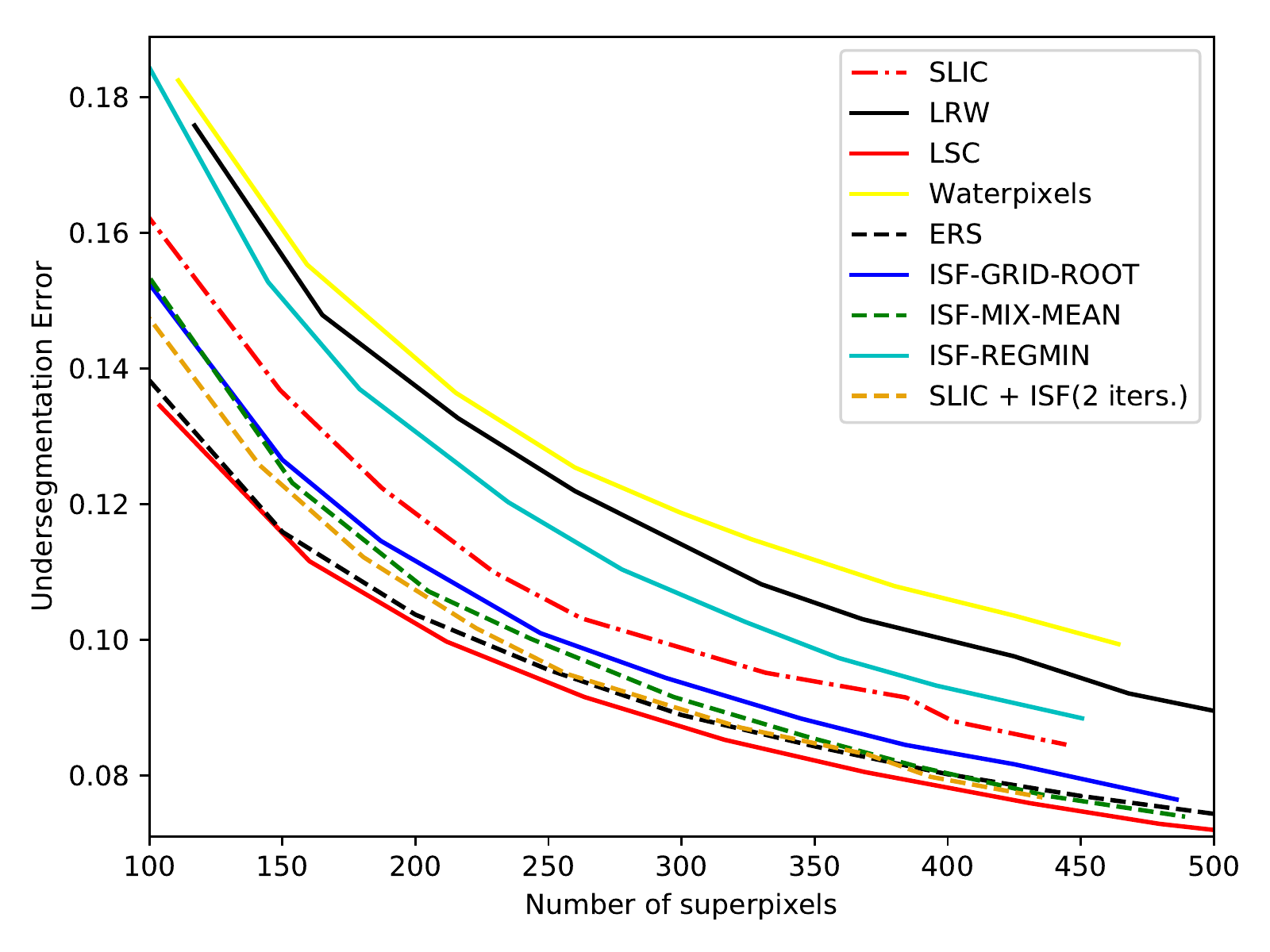} 
\end{center}
\caption{Variations of BR, UE with number of superpixels for ISF-MIX-MEAN, ISF-GRID-ROOT, ISF-REGMIN, SLIC, the combination of SLIC and ISF (two iterarions), LRW, ERS, Waterpixels and LSC methods on \textbf{Berkeley}.
We use the parameters $\alpha = 0.5$ for ISF variants, $m = 10$ (compactness parameter) for SLIC variants, $\alpha = 0.999999$ for LRW, $k = 8$ for Waterpixels and $ratio = 0.075$ for LSC.}
\label{BR_UE_vs_NS_BSD}		
\end{figure*}

\begin{figure*}[!htb]
\begin{center}

   \includegraphics[width= 7.7cm]{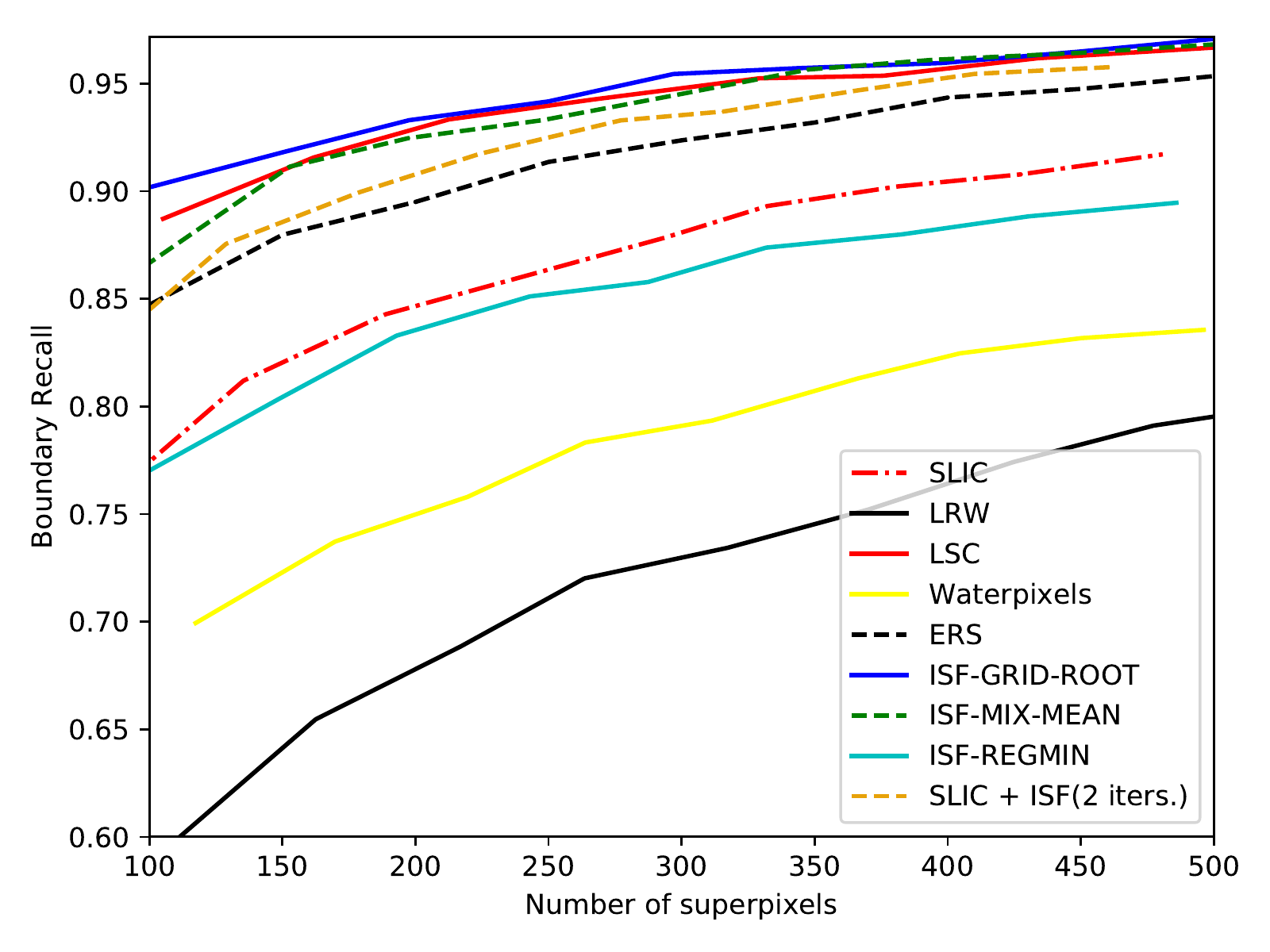} 
	 \hspace{1cm}
	 \includegraphics[width=7.7cm]{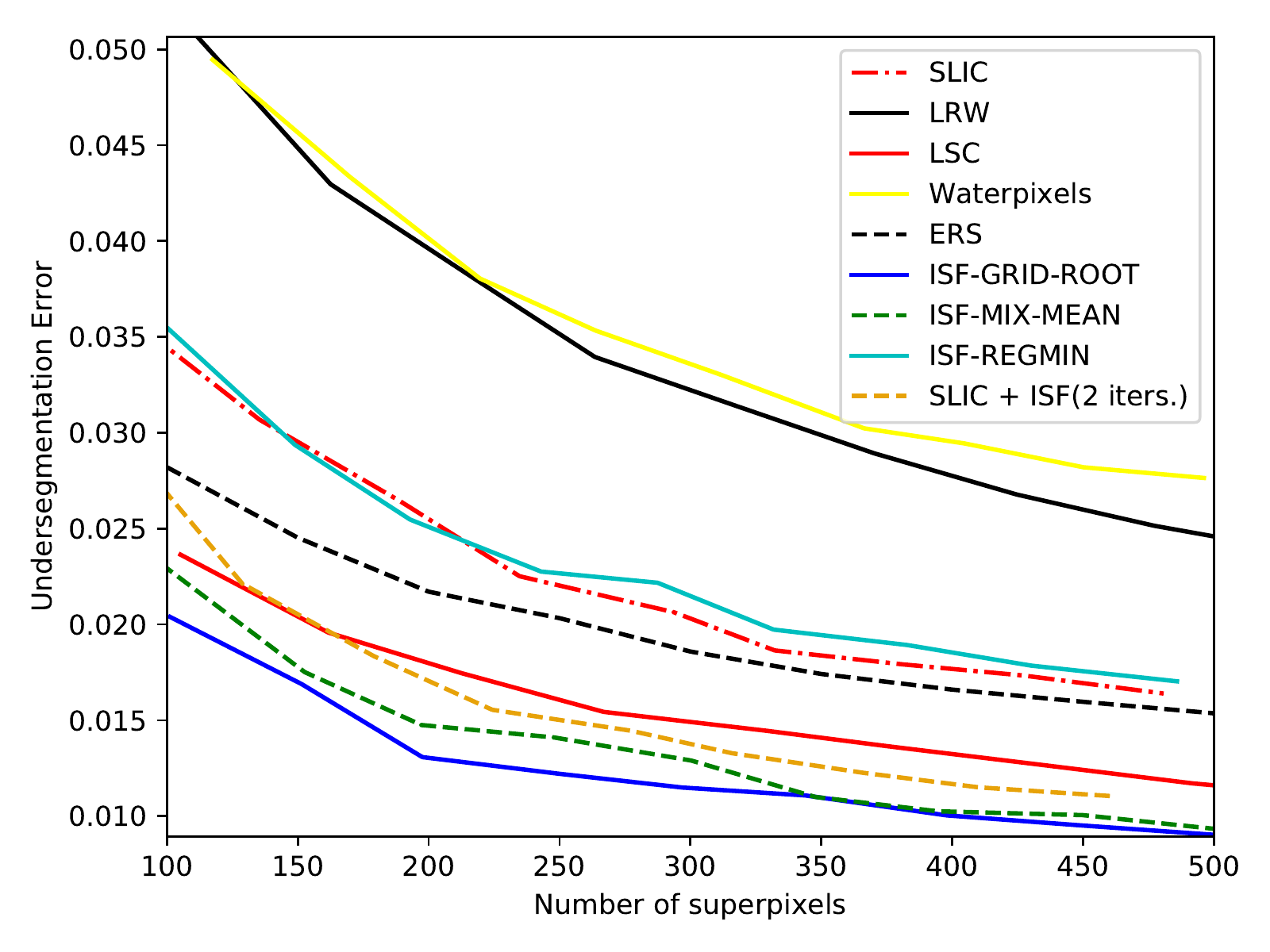} 
\end{center}
\caption{Variations of BR, UE with number of superpixels for ISF-MIX-MEAN, ISF-GRID-ROOT, ISF-REGMIN, SLIC, the combination of SLIC and ISF (two iterarions), LRW, ERS, Waterpixels and LSC methods on \textbf{Birds}.
We use the parameters $\alpha = 0.5$ for ISF variants, $m = 10$ (compactness parameter) for SLIC variants, $\alpha = 0.999999$ for LRW, $k = 8$ for Waterpixels and $ratio = 0.075$ for LSC.}
\label{BR_UE_vs_NS_Birds}		
\end{figure*}

\begin{figure*}[!htb]
\begin{center}

   \includegraphics[width= 7.7cm]{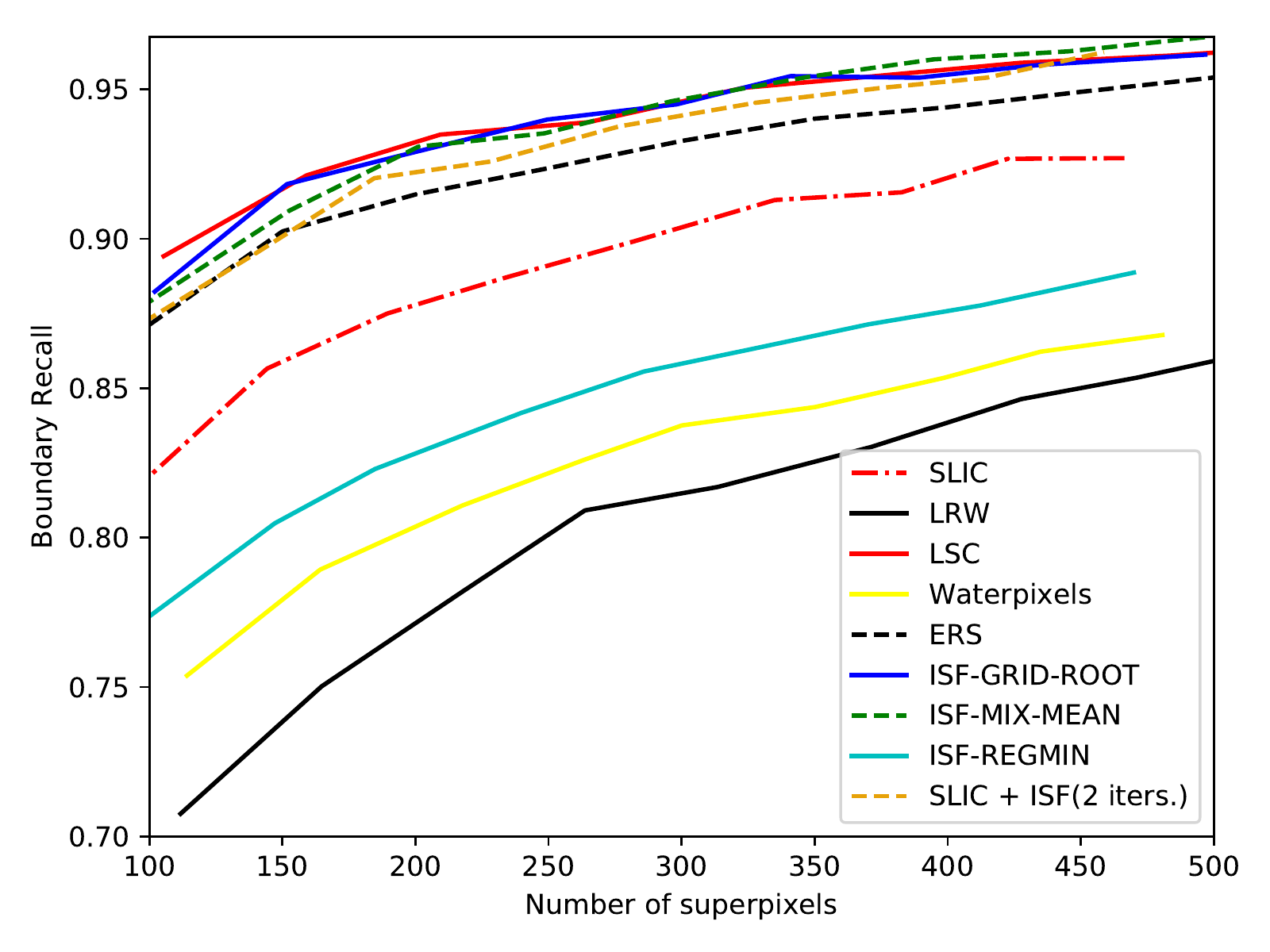} 
	 \hspace{1cm}
	 \includegraphics[width=7.7cm]{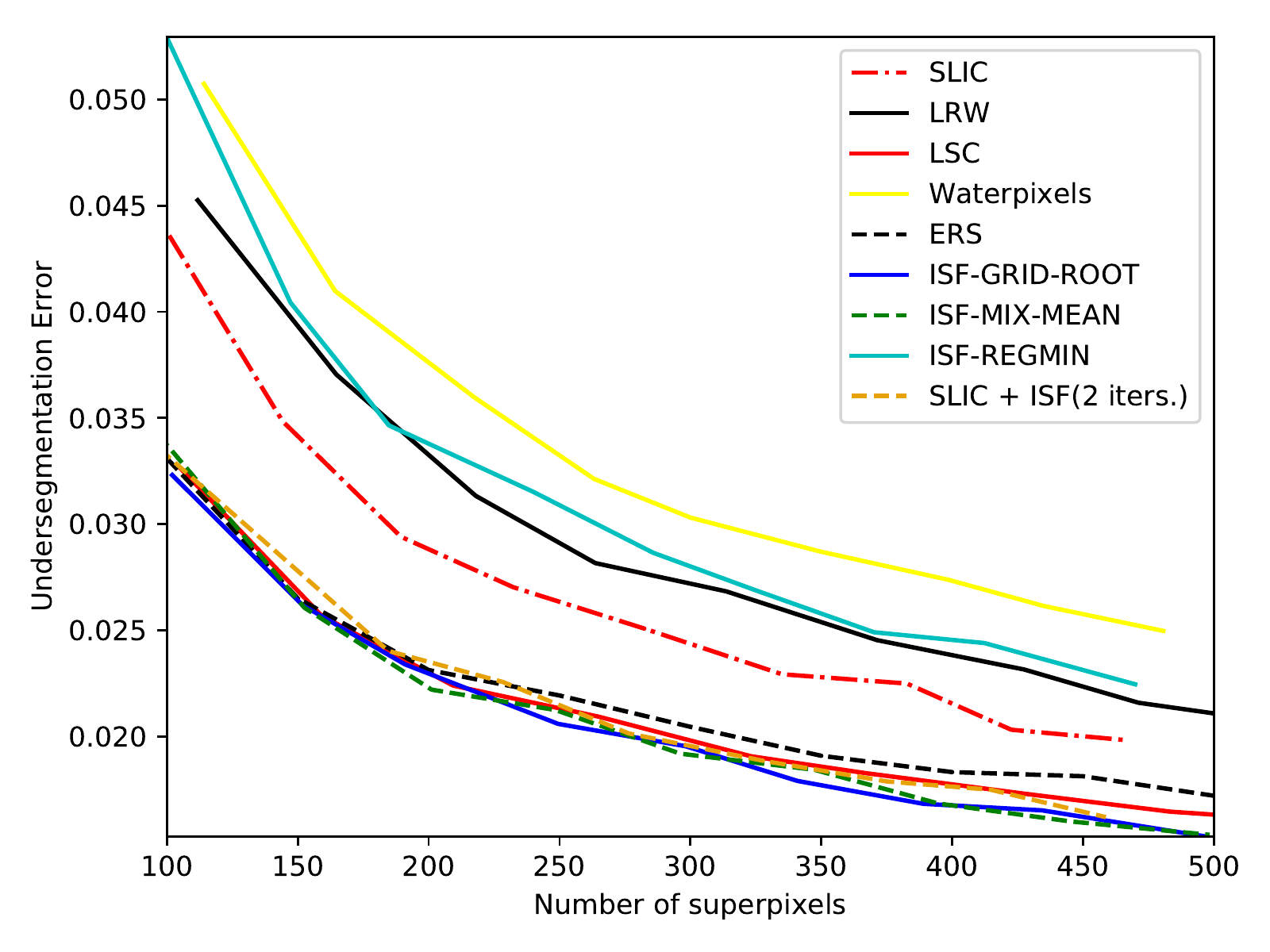} 
\end{center}
\caption{Variations of BR, UE with number of superpixels for ISF-MIX-MEAN, ISF-GRID-ROOT, ISF-REGMIN, SLIC, the combination of SLIC and ISF (two iterarions), LRW, ERS, Waterpixels and LSC methods on \textbf{Grabcut}.
We use the parameters $\alpha = 0.5$ for ISF variants, $m = 10$ (compactness parameter) for SLIC variants, $\alpha = 0.999999$ for LRW, $k = 8$ for Waterpixels and $ratio = 0.075$ for LSC.}
\label{BR_UE_vs_NS_Grabcut}		

\end{figure*}

\begin{figure*}[!htb]
\begin{center}

   \includegraphics[width= 7.7cm]{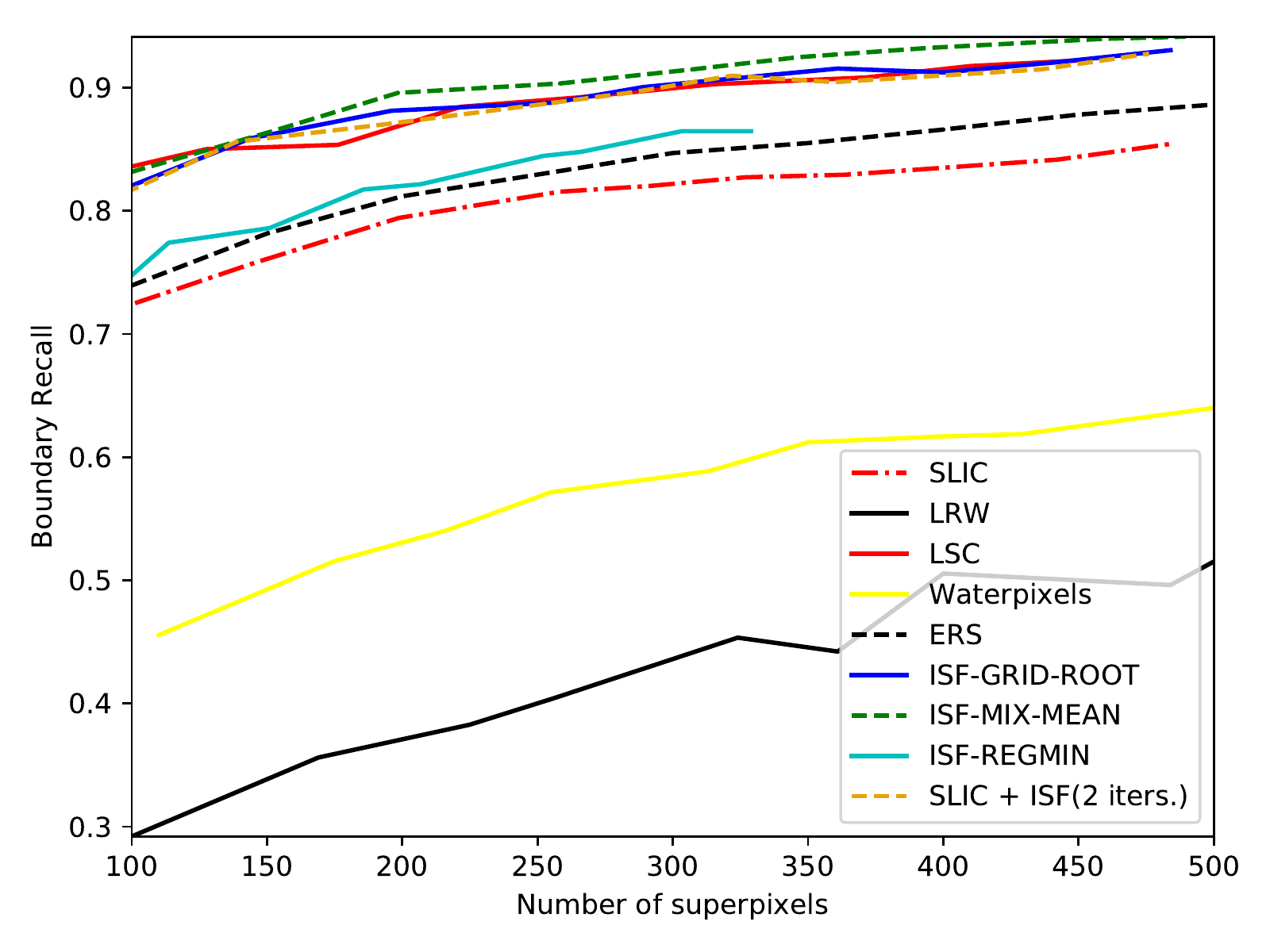} 
	 \hspace{1cm}
	 \includegraphics[width=7.7cm]{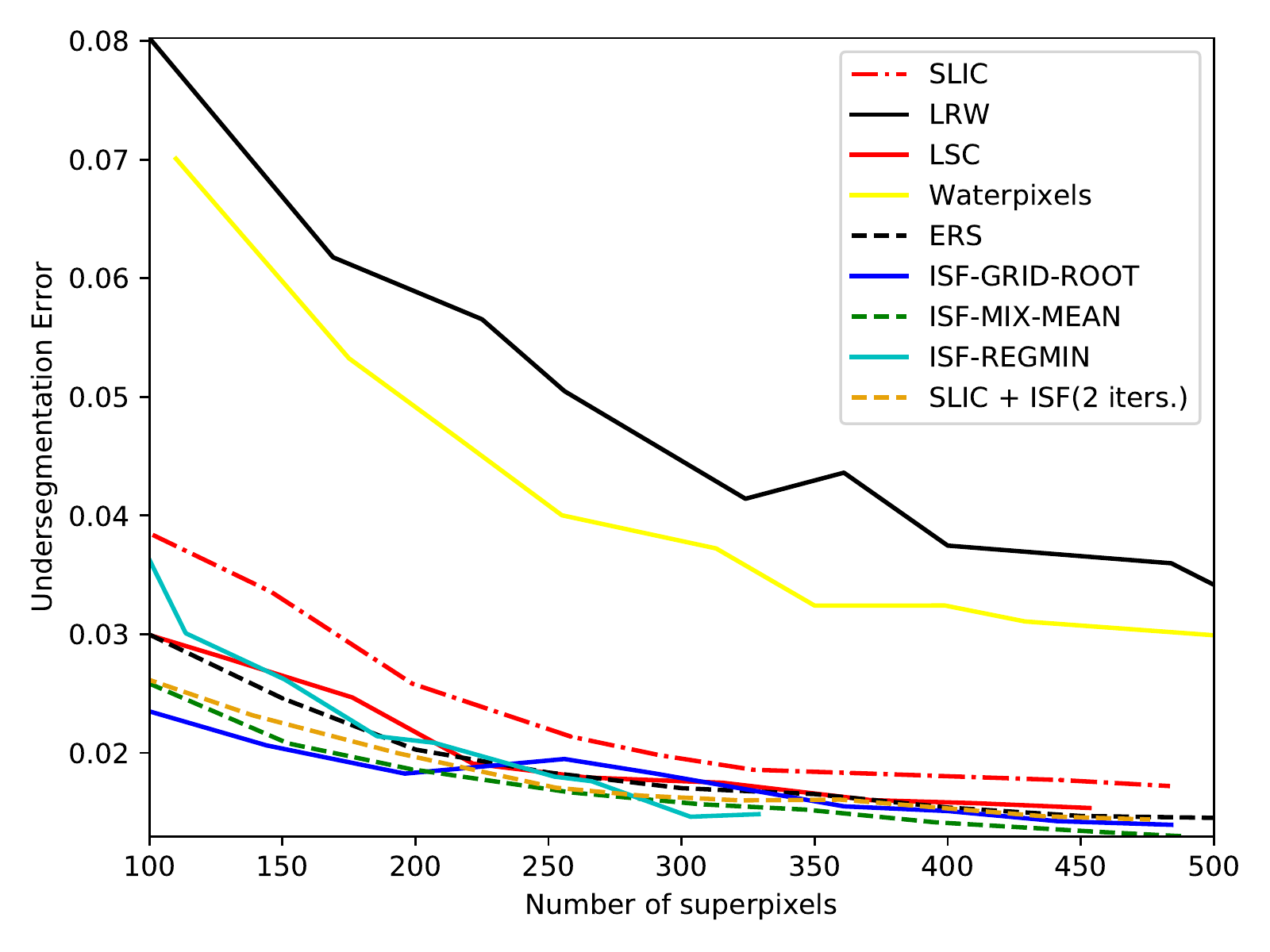} 
\end{center}
\caption{Variations of BR, UE with number of superpixels for ISF-MIX-MEAN, ISF-GRID-ROOT, ISF-REGMIN, SLIC, the combination of SLIC and ISF (two iterarions), LRW, ERS, Waterpixels and LSC methods on \textbf{Liver}.
We use the parameters $\alpha = 0.5$ for ISF variants, $m = 10$ (compactness parameter) for SLIC variants, $\alpha = 0.999999$ for LRW, $k = 8$ for Waterpixels and $ratio = 0.075$ for LSC.}
\label{BR_UE_vs_NS_Liver}		
\end{figure*}

The ISF methods are compared with five approaches from the
state-of-the-art: SLIC (Simple Linear Interative
Clustering)~\cite{Achanta2012}~\footnote{\url{http://ivrl.epfl.ch/supplementary_material/RK_SLICSuperpixels/}},
LSC (Linear Spectral
Clustering)~\cite{Chen_2017}~\footnote{\url{http://jschenthu.weebly.com/projects.html}},
ERS (Entropy Rate Superpixel)~\cite{Liu2011}, LRW (Lazy Random
Walk)~\cite{Shen2014}~\footnote{\url{https://github.com/shenjianbing/lrw14/}},
and Waterpixels~\cite{Machairas_2015}. Except for ISF-REGMIN, the
remaining ISF methods are competitive among themselves with some
differences in effectiveness. Therefore, in order to avoid busy and
confusing plots, we present the effectiveness of two of the best ISF
methods (10 iterations), ISF-GRID-ROOT and ISF-MIX-MEAN, for each
dataset. We maintain ISF-REGMIN in the plots, because it (a) uses an
integer path cost function, which allows fast computation in time
proportional to the number of pixels and independent of the number of
seeds (superpixels), (b) does not require seed recomputation, and even
being the simplest among the ISF methods, (c) it shows consistently
better effectiveness than its counterpart,
Waterpixels~\cite{Machairas_2015}. We also include a fast hybrid
approach, namely SLIC-ISF, that combines 10 iterations of SLIC for
seed estimation, followed by 2 iterations of ISF, to show that it is
competitive with the other ISF methods in most
datasets. Figures~\ref{BR_UE_vs_NS_BSD}--\ref{BR_UE_vs_NS_Liver} show
the results of this first round of experiments, using $\alpha=0.5$ and
$\beta=12$ for the ISF methods that use $f_1$ or $f_2$.

Although LSC presents the best performance (the highest BR and the
lowest UE) in Berkeley, the same is not observed in the other three
datasets. For Birds, Grabcut, and Liver, the best methods are
ISF-GRID-ROOT, ISF-GRID-ROOT (being equivalent to ISF-MIX-MEAN), and
ISF-MIX-MEAN, respectively. In Berkeley, ISFMIX-MEAN performs second
best in BR and SLIC-ISF performs second best in
UE. ISF-REGMIN is consistenly better than Waterpixels in both BR and
UE for all datasets. ERS performs well in Berkeley, but its
performance is not competitive in the other three datasets. Although
SLIC is the fastest and most used method, its performance is far from
being competitive in all datasets. Among the baselines, LSC is the
most competitive with the ISF methods. However, it seems that the
performance of LSC in UE can be negatively affected for thin and
elongated objects, such as birds. Except for Berkeley, SLIC-ISF
presents better performance than ERS in BR and UE.

\begin{figure*}[!htb]
\begin{center}
\begin{tabular}{ccc}
   \includegraphics[width=5cm]{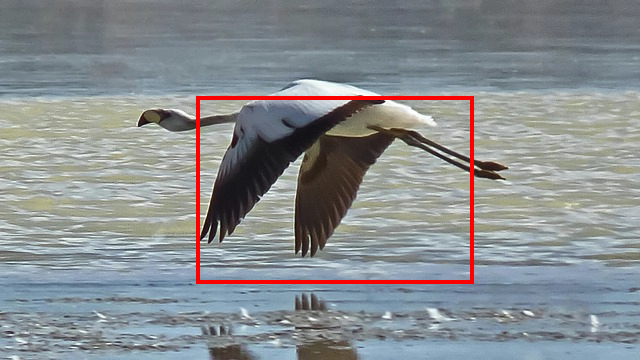} &
	 \includegraphics[width=3.35cm]{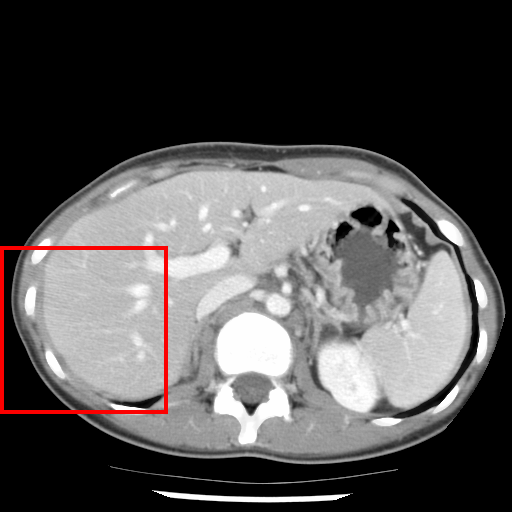} &
	 \includegraphics[width=5cm]{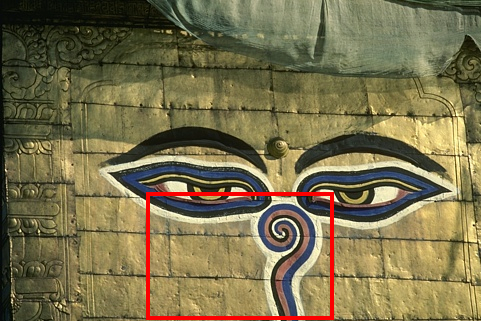}\\

    \includegraphics[width=5cm]{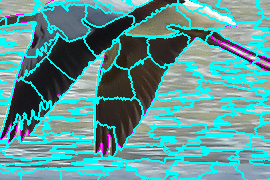} &
	 \includegraphics[width=3.35cm]{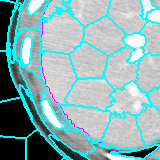} &
	 \includegraphics[width=5cm]{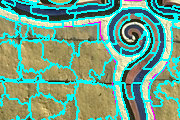}\\
   
   \includegraphics[width=5cm]{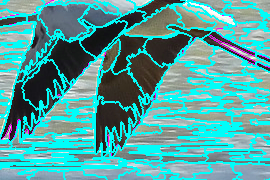} &
	 \includegraphics[width=3.35cm]{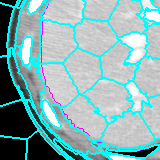} &
	 \includegraphics[width=5cm]{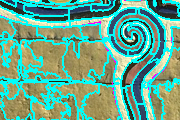} \\
   
   \includegraphics[width=5cm]{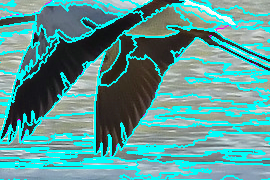} &
	 \includegraphics[width=3.35cm]{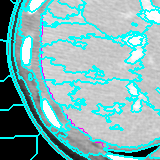} &
	 \includegraphics[width=5cm]{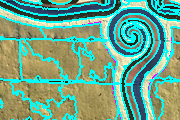}\\
   
   \includegraphics[width=5cm]{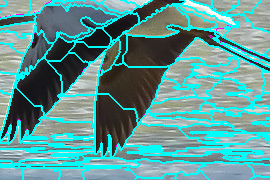} &
	 \includegraphics[width=3.35cm]{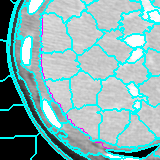} &
	 \includegraphics[width=5cm]{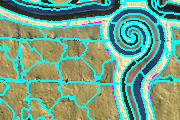}\\
	 
   (a) & (b) & (c)
\end{tabular} 
\end{center}

\caption{Examples of superpixel segmentation in (a) Birds, (b) Liver
  and (c) Berkeley, using SLIC (second row), LSC (third row), ISF with
  $\alpha = 0.5$ (fourth row), and ISF with $\alpha = 0.12$ (fifth
  row) --- ISF-GRID-ROOT (first column), ISF-MIX-MEAN (second column), and ISF-MIX-MEAN (third column). The superpixel borders are presented in cyan and the
  ground-truth borders in magenta (i.e., errors appear in magenta).}
\label{LRW_vs_ISF_vs_SLIC_Ex}
\end{figure*}

In conclusion, one cannot say that there is a winner for all datasets,
but it is clear that ISF can produce highly effective methods with
different performances depending on the dataset. In Birds, Grabcut,
and Liver, ISF shows better effectiveness than the most competitive
baseline, LSC. This shows the importance of obtaining connected
superpixels with no need for post-processing. The performance of LSC in UE is usually inferior when compared to its performance in BR. Birds dataset is clearly a case in the point. Indeed, LSC produces less regular superpixels with
high BR. In sky image segmentation, as we will see, this property of
LSC considerably impairs its effectiveness. Between ISF-GRID-ROOT and
ISF-MIX-MEAN, we can say that ISF-MIX-MEAN provides better results in
most datasets, including the application of sky image segmentation. We
believe this is related to the advantages in effectiveness of mix
sampling over grid sampling. Figure~\ref{LRW_vs_ISF_vs_SLIC_Ex} then
illustrates the quality of the segmentation in images from three
datasets using the best ISF method for the dataset, the fastest
approach, SLIC, and the most competitive baseline, LSC. Additionaly,
we show the ISF method with a choice of $\alpha=0.12$ that produces
more regular superpixels without compromising its performance in BR
and UE. This simply shows that by choice of $\alpha$, ISF can control
superpixel regularity.

\begin{figure}
  \begin{center}
    \begin{tabular}{cc}
   \includegraphics[width=4.0cm]{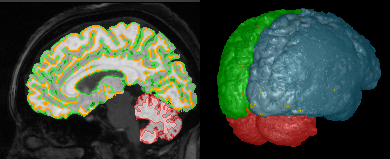} & 
   \includegraphics[width=3.0cm]{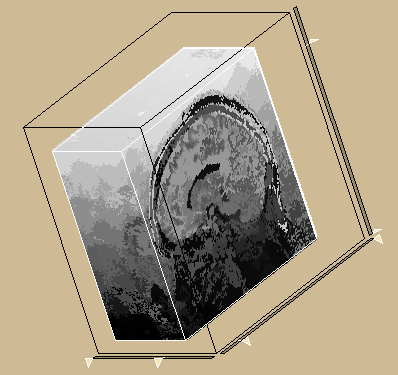} \\ 
   \includegraphics[width=3.5cm]{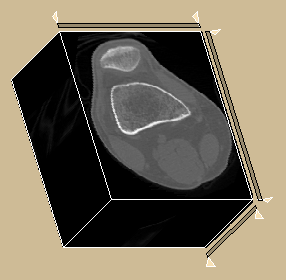} & 
   \includegraphics[width=3.5cm]{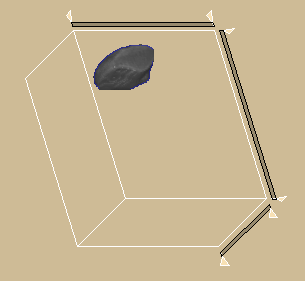}
   \end{tabular}
\end{center}
\caption{(a) Cerebellum, left and right brain hemispheres from an MR image of the brain. (b) Resulting supervoxels for one MR image of the brain. (c) CT image of a knee. (d) For a segmentation of 10 supervoxels, the patella bone is obtained as one of them.}
\label{3DImages}
\end{figure}

Second, given that the 3D extension of ISF simply requires a different
choice of adjacency relation, we present a comparison between the best
ISF method for this application (ISF-GRID-MEAN with $\alpha=0.1$), the
only baseline with 3D implementation (SLIC), and the hybrid approach
(SLIC-ISF) on volumetric MR images of the brain. In this dataset,
there are three objects of interest: cerebellum, left and right brain
hemispheres (Figure~\ref{3DImages}a). Segmentation creates supervoxels
as shown in Figure~\ref{3DImages}b. Supervoxels with more than 50\% of
their voxels inside a particular object are labeled as belonging to
that object, otherwise they are considered as part of the background
or other objects. Effectiveness is measured by f-score for three
supervoxel resolutions, given the usual image sizes: low ($N=1000$),
medium ($N=5000$), and high
($N=10000$). Table~\ref{results_3d_medical_images} shows the results
of this experiment, using a 64 bit, Core(TM) i7-3770K Intel(R) PC with
CPU speed of \@ 3.50GHz. It is not a surprise that ISF outperforms
SLIC in effectiveness. However, SLIC is exploiting parallel
computing~\footnote{Without parallel computing, SLIC would take from
  19s-23s of processing time for $N=1000$ to $N=10000$ supervoxels.}
and given that SLIC-ISF is twice faster than ISF, their equivalence in
performance above medium superpixel resolution is an excellent
result. Another interesting observation is that ISF performs better
for a value of $\alpha$ ($\alpha=0.1$) lower than $0.5$ (i.e., more
regular supervoxels).

\begin{table*}[!htb]
\renewcommand{\arraystretch}{1.5}
\caption{Fscore (mean +/- std. deviation) for cerebellum, left and right brain hemispheres in 3D MR images.}
\label{results_3d_medical_images}
\centering

\begin{tabular}{|l|c|r|r|c|r|r|c|r|r|}
\hline
 &\multicolumn{3}{c|}{$N = 1000$} &\multicolumn{3}{c|}{$N = 5000$} & \multicolumn{3}{c|}{$N = 10000$} \\
\cline{2-10}
Method &\multicolumn{1}{c|}{FScore}&\multicolumn{1}{c|}{Stdev}&\multicolumn{1}{c|}{Time(sec)}
 &\multicolumn{1}{c|}{FScore}&\multicolumn{1}{c|}{Stdev}&\multicolumn{1}{c|}{Time(sec)}
 &\multicolumn{1}{c|}{FScore}&\multicolumn{1}{c|}{Stdev}&\multicolumn{1}{c|}{Time(sec)}\\
\hline
SLIC  		              &  0.8584 &  0.0110 & 6.1 & 0.9194  & 0.0075 & 7.0 & 0.9369 & 0.0039 & 7.2\\
\hline
ISF-GRID-MEAN		      & 0.8815 & 0.0129 & 31.8 &  0.9321  & 0.0069 & 30.3 & 0.9459 & 0.0051 & 29.9\\
\hline
SLIC + ISF (two iterations)   & 0.8686 &  0.0138 & 17.3 & 0.9305 & 0.0072  & 18.0 & 0.9444 & 0.0044 & 18.0\\
\hline
\end{tabular}
\end{table*}

Figures~\ref{3DImages}c-d show another example using ISF-GRID-MEAN,
where the specification of 10 supervoxels using $\alpha=0.5$ segments
the patella bone as one of the supervoxels.

\subsection{Effectiveness in a high level application}

\begin{figure}
\begin{center}
   \includegraphics[width= 8cm]{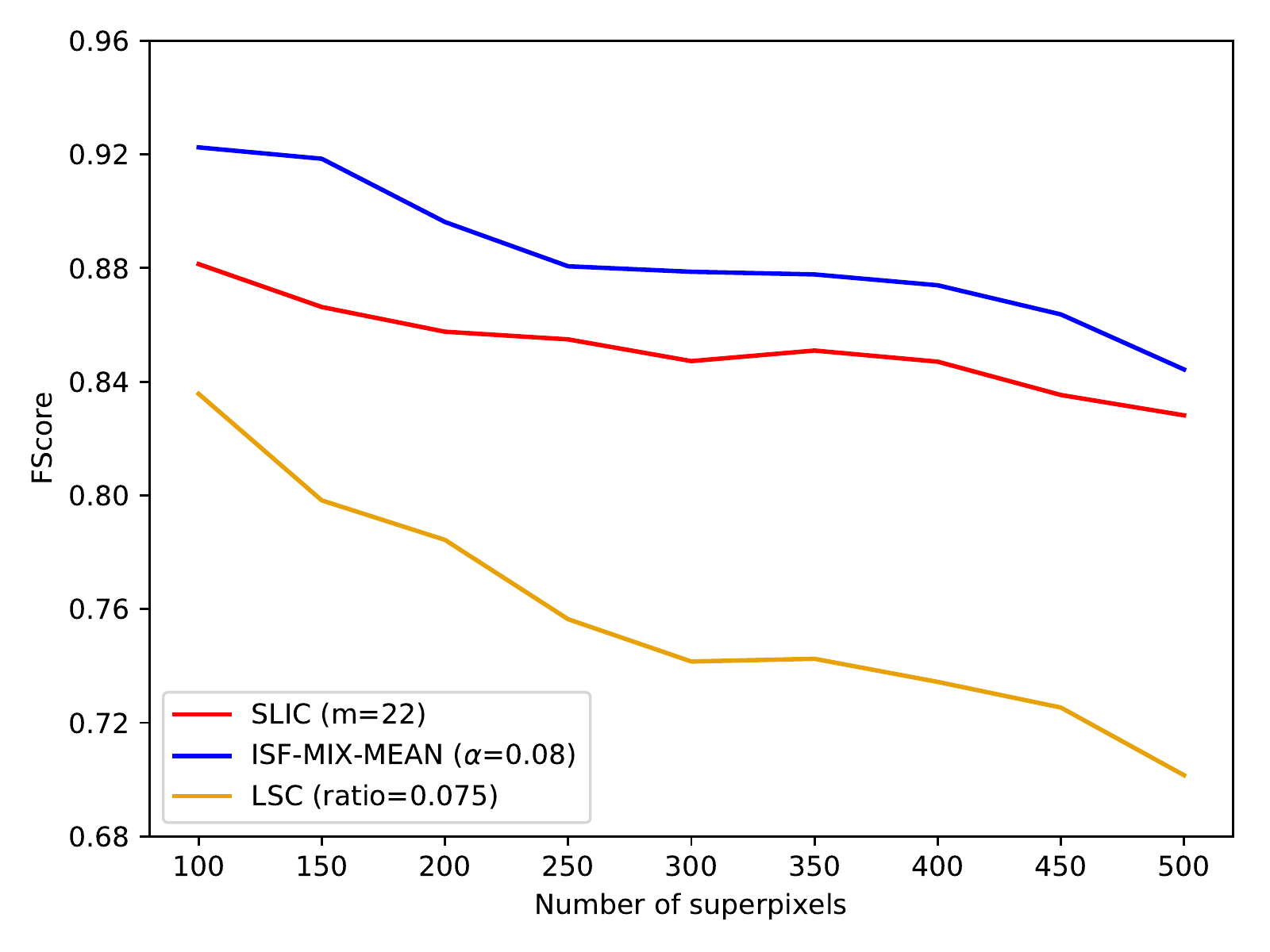}
\end{center}
\caption{Performance in f-score (Dice) for sky image segmentation:
  ISF-MIX-MEAN, SLIC, and LSC. Each method uses its best parameter
  values.}
\label{Skyseg_FSCORE_vs_NS}
\end{figure}

When considering a high level application, such as superpixel-based
image segmentation, the label assigment to superpixels follows some
independent and automatic rule. In this section, we evaluate the
performance of the best ISF method (ISF-MIX-MEAN) in this application,
namely \emph{sky image segmentation}, in comparison with the fastest
method (SLIC) and the most competitive baseline (LSC). We use a simple
yet effective sky segmentation algorithm, as presented
in~\cite{SkySegmentation}. This algorithm uses the mean color of the
superpixels and a threshold defined in the Lab color space to merge
superpixels. The region (set of superpixels) in the top of the image
that contains the larger number of pixels is selected as the sky
region. Figure~\ref{Skyseg_FSCORE_vs_NS} shows the results of f-score
for this experiment for varying number of superpixels. Again, ISF with
$\alpha = 0.08$ (more regular superpixels) performs better than the
others.

The use of lower values of $\alpha$ in the segmentation of 3D MR
images of the brain and in this application strongly suggests that
superpixel regularity has some importance as well as boundary
adherence. It is also interesting to observe that SLIC outperforms LSC
in this application.

\subsection{Efficiency}
\label{efficiency}

SLIC is acknowledged as one of the fastest superpixel segmentation
methods~\cite{Neubert2012}. In this section, we compare the processing
times in one of the datasets (Berkeley) for the ISF methods used in
the effectiveness experiments in 2D, using different superpixel
resolutions and values of the parameter $\alpha$, SLIC, LSC, and ERS
(the two most competitive methods in Berkeley). Table
\ref{table_proc_time} shows the average processing time in seconds of
the methods, without taking into account the I/O operations and
pre-processing (e.g. RGB to Lab conversion), and using the same
machine specification for Table \ref{results_3d_medical_images}. Note
that the optimized code of ISF can run faster with higher number of
superpixels and lower value of $\alpha$ (more regular
superpixels). This can be explained by the use of the differential
image foresting transform~\cite{Condori2017}, whose processing time is
$O(N\log N)$ where $N$ is the number of pixels in the modified regions
of the image. As the number of superpixels increases and their shapes
become more compact, the sizes of the modified regions per iteration
reduce. Note that ISF can be more efficient than LSC and ERS in
general, and depending on the choices of $\alpha$ and number of
superpixels, ISF can achieve processing time competitive with SLIC.

\begin{table*}[!htb]
\renewcommand{\arraystretch}{1.5}
\caption{Average processing time for superpixel segmentation in the Berkeley dataset.}
\label{table_proc_time}
\centering

\begin{tabular}{|l|c|c|c|c|}
\hline
 &\multicolumn{1}{c|}{$N = 250$} &\multicolumn{1}{c|}{$N = 500$} & \multicolumn{1}{c|}{$N = 1000$} & \multicolumn{1}{c|}{$N = 5000$} \\
\cline{2-5}
Method &\multicolumn{1}{c|}{Time (sec)}
 &\multicolumn{1}{c|}{Time (sec)}
 &\multicolumn{1}{c|}{Time (sec)}
 &\multicolumn{1}{c|}{Time (sec)} \\
\hline 
ISF-MIX-MEAN ($\alpha = 0.5$) & 0.248 & 0.227 & 0.199 & 0.127 \\
\hline
ISF-MIX-MEAN ($\alpha = 0.12$) & 0.158 & 0.129 & 0.101 & 0.067 \\
\hline
ISF-MIX-MEAN ($\alpha = 0.04$) & 0.075 & 0.066 & 0.057 & \textbf{0.049} \\
\hline
ISF-GRID-ROOT ($\alpha = 0.5$) & 0.250 & 0.249 & 0.243 & 0.201 \\
\hline
ISF-GRID-ROOT ($\alpha = 0.12$) & 0.257 & 0.253 & 0.236 & 0.159 \\
\hline
ISF-GRID-ROOT ($\alpha = 0.04$) & 0.244 & 0.235 & 0.210 & 0.127 \\
\hline
SLIC  & \textbf{0.036}  & 0.038 & 0.041 & 0.042 \\
\hline
SLIC + ISF (two iterations) & 0.104 & 0.105 & 0.108 & 0.109\\
\hline
ISF-REGMIN                & 0.055 & 0.056 & 0.057 &  0.057\\
\hline 
LSC                      & 0.257 & 0.259 & 0.262 & 0.267 \\
\hline
ERS                      & 0.952 &  1.012 & 1.065 & 1.224 \\
\hline
\end{tabular}
\end{table*}

\section{Conclusion}
\label{CFW}

We present an iterative spanning forest (ISF) framework, based on
sequences of image foresting transforms (IFTs) for the generation of
superpixels based on different choices of seed sampling strategies,
connectivity functions, adjacency relations, and seed recomputation
strategies. We also introduce a new seed sampling strategy, which can
provide better results than grid sampling for most datasets, and new
connectivity functions for the IFT framework.

In the supplementary material, we prove that ISF converges and outputs
connected superpixels --- a property that avoids the post-processing
step required in several other approaches. We also demonstrate by
extensive experiments that the ISF superpixels can be computed fast
with high value of boundary recall and low value of undersegmentation
error, making ISF competitive or superior to several state-of-the-art
methods in effectiveness and efficiency.

As shown, the compromise between boundary adherence and superpixel
regularity in ISF can be controlled by choice of the parameter
$\alpha$ in Equations~\ref{eq_fd1} and~\ref{eq_fd2}. Indeed, more
superpixel regularity has shown to be important for sky image
segmentation and 3D MR image segmentation of the brain. This result
requires further and more careful investigation. We also plan to
develop a hierarchical iterative spanning forest (HISF) framework by
applying the IFT algorithm recursively on the resulting region
adjacency graphs of ISF at each level of the hierarchy. We believe
that HISF can further speed-up IFT-based superpixel/supervoxel
segmentation and improve high level applications, such as active
learning for superpixel classification.

\section{Supplementary Material}
\label{supplementary_material}

\subsection{ISF Theoretical Properties}
\label{properties}

Now, we will discuss some theoretical properties of ISF.
Let ${\cal P}^i_1$, ${\cal P}^i_2$, $\ldots$, ${\cal P}^i_k$
be the image partition (i.e., $\bigcup_{j=1}^k {\cal P}^i_j = {\cal I}$)
into $k$ superpixels obtained at the $i^{th}$ iteration by
the seeds $s^i_1$, $s^i_2$, $\ldots$, $s^i_k$.
Let's also consider the following definitions:

\begin{sloppypar}
\begin{defi}[Constrained Paths]
Let $B$ denote a subset of the vertex set of $G$.
A path $\pi {\scriptstyle s \overset{B}{\leadsto} t} = \langle t_1 = s,t_2,\ldots,t_n = t \rangle$
indicates a path which is constrained inside the subgraph induced by $B$
(i.e., $t_i \in B$, $i=1,\ldots,n$ and $(t_i,t_{i+1})\in {\cal A}$, $i=1,\ldots,n-1$).
\label{d.cp}
\end{defi}
\end{sloppypar}

\begin{sloppypar}
\begin{defi}[Optimum-Constrained Paths]
A path $\pi {\scriptstyle s \overset{B}{\leadsto} t}$ is \emph{optimum-constrained} in $B$ if $f(\pi {\scriptstyle s \overset{B}{\leadsto} t}) \leq f(\tau {\scriptstyle x \overset{B}{\leadsto} t})$ for any other constrained path $\tau {\scriptstyle x \overset{B}{\leadsto} t}$ in $B$ with the same destination node $t$.
The notation $\overset{\ast}{\pi} {\scriptstyle s \overset{B}{\leadsto} t}$ will be used to explicitly indicate an optimum-constrained path.
\label{d.ocp}
\end{defi}
\end{sloppypar}

Based on these definitions, we have the following propositions:
\begin{sloppypar}
\begin{prop}[Optimum-Constrained Path Trees by ISF]
The spanning tree of each superpixel ${\cal P}^i_j$ computed by ISF with $f$, $j=1,\ldots,k$, is an \emph{Optimum-Constrained Path Tree} in ${\cal P}^i_j$.
That is, the paths $\pi^P_t$ 
computed for the non-smooth connectivity function $f$
are optimum-constrained paths with respect to their superpixels ${\cal P}^i_j$ (i.e., $\pi^P_t = \overset{\ast}{\pi} {\scriptstyle s^i_j \overset{{\cal P}^i_j}{\leadsto} t}$).
\label{p.opct}
\end{prop}
\end{sloppypar}

Proposition~\ref{p.opct} can be proved by noting that each superpixel
${\cal P}^i_j$ has an unique seed $s^i_j$ and the function $f$
becomes a smooth function~\cite{Falcao2004}, in the subgraph induced
by ${\cal P}^i_j$, for this single seed.

Let the set of boundary pixels between neighboring superpixels
for each superpixel ${\cal P}^i_j$ be defined as
${\cal B}({\cal P}^i_j) = \left\{ t \in {\cal P}^i_j \mid \exists s \in {\cal A}(t) \mbox{ such that } s \notin {\cal P}^i_j \right\}$. Then, we can also have the following property:

\begin{sloppypar}
\begin{prop}[Boundary Protection]
For any pixel $t \in {\cal B}({\cal P}^i_j)$, $j=1,\ldots,k$, if
$s \in {\cal A}(t)$ is a pixel such that $s \in {\cal P}^i_l$ and $l \neq j$,
we have that $f(\overset{\ast}{\pi} {\scriptstyle s^i_j \overset{{\cal P}^i_j}{\leadsto} t}) \leq f(\overset{\ast}{\pi} {\scriptstyle s^i_l \overset{{\cal P}^i_l}{\leadsto} s} \cdot \langle s,t\rangle)$.
\label{p.bleq}
\end{prop}
\end{sloppypar}

Basically, this proposition states that each superpixel ${\cal P}^i_j$
is surrounded by boundary pixels ${\cal B}({\cal P}^i_j)$, which are
equally or more strongly connected to their seeds $s^i_j$ than to neighboring 
superpixels through any direct extension of their respective 
optimum-constrained paths. The proposition follows from the ordered 
propagation of the priority queue and from the fact that $f$ is a non-decreasing
function. We have two cases, depending on which pixel ($t$ or $s$) is
firstly removed from $Q$ in Line 14.  If $t$ is removed prior to $s$,
then we have that $f(\pi^P_t) \leq f(\pi^P_s)$, which implies
that $f(\pi^P_t) \leq f(\pi^P_s \cdot \langle s,t\rangle)$,
since $f$ is a non-decreasing function. Otherwise, if $s$ is
removed before $t$ from $Q$, we have that $f(\pi^P_s \cdot \langle
s,t\rangle)$ is surely evaluated in Line 17, since $S(t)\neq Black$.
So $f(\pi^P_t)$ cannot be worse than $f(\pi^P_s \cdot \langle
s,t\rangle)$, otherwise node $t$ would have been conquered by the path
$\pi^P_s \cdot \langle s,t\rangle$. Therefore, in both cases, we have
$f(\pi^P_t = \overset{\ast}{\pi} {\scriptstyle s^i_j
  \overset{{\cal P}^i_j}{\leadsto} t}) \leq f(\pi^P_s \cdot
\langle s,t\rangle = \overset{\ast}{\pi} {\scriptstyle s^i_l
  \overset{{\cal P}^i_l}{\leadsto} s} \cdot \langle s,t\rangle)$.

Now, we state our two theorems with proofs given in the next sections:

\begin{sloppypar}
\begin{theo}[Connectedness Theorem]
\label{connth}

ISF with above choices of connectivity functions, sampling strategies
and adjacency relation guarantees generation of connected superpixels.

\end{theo}
\end{sloppypar}

\begin{sloppypar}
\begin{theo}[Convergence Theorem]
\label{convth}
If the new seeds $s^{i+1}_j$ for the next iteration $i+1$,
are selected such that $\sum_{t \in {\cal P}^i_j} f(\overset{\ast}{\pi} {\scriptstyle s^{i+1}_j \overset{{\cal P}^i_j}{\leadsto} t}) < \sum_{t \in {\cal P}^i_j} f(\overset{\ast}{\pi} {\scriptstyle s^i_j \overset{{\cal P}^i_j}{\leadsto} t})$, $j=1,\ldots,k$ and if $f$ is a
smooth function, then the $ISF$ algorithm is guaranteed to converge.
\label{t.ct}
\end{theo}
\end{sloppypar}

\subsection{Proof of Theorem \ref{connth}}
Let's prove that, at the end of any iteration of the main loop (Line 2),
the image partition computed by the label map $L_s$
results in a set of connected superpixels.
Let ${\cal P}^i_1$, ${\cal P}^i_2$, $\ldots$, ${\cal P}^i_k$
be the image partition into $k$ superpixels obtained
at the end of the $ith$ iteration by the seeds $s^i_1$, $s^i_2$, $\ldots$, $s^i_k$.
The superpixels ${\cal P}^i_j$, $j=1,\ldots,k$, are gradually computed,
in the loop of Lines 13-25,
by the successive removal of pixels from $Q$ (Lines 14-15), such that at
any instant ${\cal P}^i_j = \left\{ t \in {\cal I} \mid L_s(t) = j \mbox{ and } S(t) = Black\right\}$.

The generation of connected superpixels ${\cal P}^i_j$ can be proved by mathematical induction. 
In the base case, we have initially each superpixel ${\cal P}^i_j$ being composed
exclusively by its corresponding seed $s^i_j$, which is obviously connected.
Note that the seeds are initialized with the lowest possible cost (Lines 7-12),
and thus are the first pixels to leave the priority queue $Q$.

The condition $S(t) \neq Black$ in Line 16 guarantees that any pixel
$t$ in ${\cal P}^i_j$ cannot be later removed from ${\cal P}^i_j$ and
added to another superpixel, since changes in the labelling $L_s(t)$
can only occur at Line 20.  So in the inductive step, we have only to
prove that the connectedness of ${\cal P}^i_j$, $j=1,\ldots,k$, is
preserved when a new node $s$ is added to ${\cal P}^i_j$, after it
gets removed from $Q$ in Lines 14-15.  According to Lines 19-20, the
predecessor $P(s)$ of node $s$ has its same label, i.e., $L_s(P(s)) =
L_s(s)$.  Therefore, node $s$ is necessarily connected to a superpixel
${\cal P}^i_j$, where $j = L_s(s)$.  The 4-neighborhood
guarantees connected superpixels not only in the graph topology, but
also in the image domain.  This symmetric adjacency leads to a
strongly connected digraph ensuring that all pixels are assigned to
some superpixel.  So, with the given choice of connectivity and
adjacency, we guarantee generation of an image partition into
connected superpixels.

\subsection{Proof of Theorem \ref{convth}}

\begin{proof}
For each iteration $i$, consider the functional $F_i$:
\begin{equation}
F_i = \sum_{j=1}^{k} \sum_{t \in {\cal P}^i_j} f(\overset{\ast}{\pi} {\scriptstyle s^i_j \overset{{\cal P}^i_j}{\leadsto} t}) = \sum_{t \in {\cal I}} C_i(t) \label{e.func},
\end{equation}
where $C_i(t)$ denotes the connectivity map $C$ computed by the IFT at
its $ith$ execution.

For the new considered seeds, we have that:
\begin{equation}
F_i = \sum_{j=1}^{k} \sum_{t \in {\cal P}^i_j} f(\overset{\ast}{\pi} {\scriptstyle s^i_j \overset{{\cal P}^i_j}{\leadsto} t}) > \sum_{j=1}^{k} \sum_{t \in {\cal P}^i_j} f(\overset{\ast}{\pi} {\scriptstyle s^{i+1}_j \overset{{\cal P}^i_j}{\leadsto} t}). \label{e.func_new}
\end{equation}

The superpixels ${\cal P}^{i+1}_j$ computed in the next iteration
are usually different from the previous ${\cal P}^i_j$, but ${\cal P}^{i+1}_j \cap {\cal P}^i_j \neq \emptyset$ because of the seed imposition and $s^{i+1}_j \in {\cal P}^i_j$.
For any pixel $p \in {\cal P}^i_j$ such that $p \notin {\cal P}^{i+1}_j$,
if $f$ is a smooth function we may conclude that $p$ was conquered,
in the iteration $i+1$, by an
optimum path $\overset{\ast}{\pi} {\scriptstyle s^{i+1}_l \overset{{\cal P}^{i+1}_l}{\leadsto} p}$,
such that $l \neq j$ and
$f(\overset{\ast}{\pi} {\scriptstyle s^{i+1}_l \overset{{\cal P}^{i+1}_l}{\leadsto} p}) \leq f(\overset{\ast}{\pi} {\scriptstyle s^{i+1}_j \overset{{\cal P}^i_j}{\leadsto} p})$.
So we have that:

\begin{equation}
\sum_{j=1}^{k} \sum_{t \in {\cal P}^i_j} f(\overset{\ast}{\pi} {\scriptstyle s^{i+1}_j \overset{{\cal P}^i_j}{\leadsto} t}) \geq\sum_{j=1}^{k} \sum_{t \in {\cal P}^{i+1}_j} f(\overset{\ast}{\pi} {\scriptstyle s^{i+1}_j \overset{{\cal P}^{i+1}_j}{\leadsto} t}) = F_{i+1}. \label{e.func_smooth}
\end{equation}

By combining Equations~\ref{e.func_smooth} and~\ref{e.func_new} we have:

\begin{equation}
F_i = \sum_{j=1}^{k} \sum_{t \in {\cal P}^i_j} f(\overset{\ast}{\pi} {\scriptstyle s^i_j \overset{{\cal P}^i_j}{\leadsto} t}) > \sum_{j=1}^{k} \sum_{t \in {\cal P}^{i+1}_j} f(\overset{\ast}{\pi} {\scriptstyle s^{i+1}_j \overset{{\cal P}^{i+1}_j}{\leadsto} t}) = F_{i+1}. \label{e.func_final}
\end{equation}

Since each iterative step necessarily lowers the
value of $F_i$ ($F_{i+1} < F_i$) and $F_i$ is lower bounded by zero (the cost of trivial paths from seeds), we have the proof of convergence.
As we increase $i$, $F_i$ will converge to a local minimum.\end{proof}

Note that if for the next iteration $i+1$, the best seed leads to
$\sum_{t \in {\cal P}^i_j} f(\overset{\ast}{\pi} {\scriptstyle s^{i+1}_j \overset{{\cal P}^i_j}{\leadsto} t}) = \sum_{t \in {\cal P}^i_j} f(\overset{\ast}{\pi} {\scriptstyle s^i_j \overset{{\cal P}^i_j}{\leadsto} t})$, then we should select the same
seed (i.e., $s^{i+1}_j = s^i_j$) in order to stabilize the results.

Next, we discuss the convergence for the case of the non-smooth function $f$.

In the update step, for each superpixel ${\cal P}^i_j$,
we select a well centralized pixel $s^{i+1}_j \in {\cal P}^i_j$
with a color closer to its mean color.
Since $f$ is an additive function, a central position will usually lower $\sum_{t \in {\cal P}^i_j} f(\overset{\ast}{\pi} {\scriptstyle s^{i+1}_j \overset{{\cal P}^i_j}{\leadsto} t})$ by reducing the length of the computed paths, and the usage of a color closer to the mean color will reduce the cost of $\|I(t)-I_r \|$ in the computation of $f$.

The problem with the usage of the non-smooth function $f$ is that we
can no longer guarantee the validity of Equation~\ref{e.func_smooth}.
That is, for a pixel $p \in {\cal P}^i_j$ such that $p \notin {\cal
  P}^{i+1}_j$, $p$ may be conquered by a path $\overset{\ast}{\pi}
{\scriptstyle s^{i+1}_l \overset{{\cal P}^{i+1}_l}{\leadsto} p}$, such
that $l \neq j$ and $f(\overset{\ast}{\pi} {\scriptstyle s^{i+1}_l
  \overset{{\cal P}^{i+1}_l}{\leadsto} p}) > f(\overset{\ast}{\pi}
{\scriptstyle s^{i+1}_j \overset{{\cal P}^i_j}{\leadsto} p})$.  One
possible way to handle this problem is by detecting the above
situation on-the-fly and by adding new dummy seeds in these
regions. By adding more seeds the function $f$ always converges to a
smooth function.

Note also that $F_i$ decreases as we add more seeds. The dummy seeds
can later be promoted to real seeds and generate their own
superpixels, or can be eliminated after the convergence, leaving their
regions to be conquered by their neighboring superpixels at a last IFT
execution.


\section*{Acknowledgment}
Ananda S. Chowdhury was supported through a FAPESP Visiting Scientist
Fellowship under the grant $2015/01186-6$. The authors thank the
financial support of CAPES, CNPq (grants $479070/2013-0$,
$302970/2014-2$ and $308985/2015-0$) and FAPESP (grants
$2014/12236-1$, $2016/14760-5$ and $2014/12236-1$). We thank
Dr. J.K. Udupa (MIPG-UPENN) for the CT images of the liver and
Dr. F. Cendes (FCM-UNICAMP) for the MR images of the brain.

\ifCLASSOPTIONcaptionsoff
  \newpage
\fi

\bibliographystyle{IEEEtran}
\bibliography{isf}
\vspace{-6 mm}
\begin{IEEEbiographynophoto}{John E. Vargas Mu{\~{n}}oz}
John E. Vargas Mu{\~{n}}oz received the B.Sc. degree in informatics 
engineering from the National University of San Antonio Abad in Cusco, 
Cusco, Peru, in 2010, and the master's degree in computer science 
from the University of Campinas, Campinas, Brazil, in 2015.
He is currently pursuing the Ph.D. degree with the University of Campinas.
His research interests include machine learning, image segmentation and 
remote sensing image classification.
\end{IEEEbiographynophoto}

\vfill
\begin{IEEEbiographynophoto}{Ananda S. Chowdhury}
Ananda S. Chowdhury earned his Ph.D. in Computer Science from the University of Georgia, Athens, Georgia in July 2007. From August 2007 to December 2008, he worked as a postdoctoral fellow in the department of Radiology and Imaging Sciences at the National Institutes of Health, Bethesda, Maryland. At present, he is working as an Associate Professor in the department of Electronics and Telecommunication Engineering at Jadavpur University, Kolkata, India where he leads the Imaging Vision and Pattern Recognition group. He has authored or coauthored more than forty-five papers in leading international journals and conferences, in addition to a monograph in the Springer Advances in Computer Vision and Pattern Recognition Series. His research interests
include computer vision, pattern recognition, biomedical image processing, and multimedia analysis.
Dr. Chowdhury is a senior member of the IEEE and the IAPR TC member of Graph-Based Representations in Pattern Recognition. He currently serves as an Associate Editor of Pattern Recognition Letters and his Erd{\"o}s number is 2.
\end{IEEEbiographynophoto}

\begin{IEEEbiographynophoto}{Eduardo B. Alexandre}
  Eduardo B. Alexandre is currently pursuing Masters in Computer Science
  at the Institute of Mathematics and Statistics (IME) of the University
  of S\~{a}o Paulo (USP). He graduated in Digital Games by University of Vale
  do Itaja\'{i} (UNIVALI). Works in Computer Vision and Digital Image Processing
  areas, with emphasis in image segmentation. He worked in LAPIX, associated
  lab of National Research Institute on Digital Convergence (INCoD), of Federal
  University of Santa Catarina (UFSC) and 4Vision Lab of University do Vale
  do Itaja\'{i} (UNIVALI).
\end{IEEEbiographynophoto}

\begin{IEEEbiographynophoto}{Felipe L. Galv\~{a}o}
Felipe L. Galv\~{a}o  is currently pursuing a M.Sc. degree in Computer Science at 
University of Campinas (UNICAMP), SP, Brazil. He received a B.Sc in Computer 
Engineering from the University of Campinas (UNICAMP) in 2017. His research 
interests include machine learning and image processing, with emphasis on 
active learning and image segmentation.
\end{IEEEbiographynophoto}

\begin{IEEEbiographynophoto}{Paulo A. V. Miranda}
Dr. Paulo A. V. Miranda is currently professor at the Institute of Mathematics 
and Statistics (IME) of the University of S\~{a}o Paulo (USP), SP, Brazil. 
He received a B.Sc. in Computer Engineering (2003) and a M.Sc. in 
Computer Science (2006) from the University of Campinas (UNICAMP), SP, Brazil. 
During 2008-2009, he was with the Medical Image Processing Group, 
Department of Radiology, University of Pennsylvania, Philadelphia, USA, 
where he worked on image segmentation for his doctorate. He got his 
doctorate in Computer Science from the University of Campinas (UNICAMP) 
in 2009. After that, he worked as a post-doctoral researcher in a project in 
conjunction with the professors of the Department of Neurology, Unicamp. 
He has experience in computer science, with emphasis on computer vision, 
image processing and pattern recognition.
\end{IEEEbiographynophoto}

\begin{IEEEbiographynophoto}{Alexandre X. Falc\~{a}o}
Alexandre Xavier Falc\~{a}o is full professor at the Institute of Computing, University of Campinas, Campinas, SP, Brazil. He received a B.Sc. in Electrical Engineering from the Federal University of Pernambuco, Recife, PE, Brazil, in 1988. He has worked in biomedical image processing, visualization and analysis since 1991. In 1993, he received a M.Sc. in Electrical Engineering from the University of Campinas, Campinas, SP, Brazil. During 1994-1996, he worked with the Medical Image Processing Group at the Department of Radiology, University of Pennsylvania, PA, USA, on interactive image segmentation for his doctorate. He got his doctorate in Electrical Engineering from the University of Campinas in 1996. In 1997, he worked in a project for Globo TV at a research center, CPqD-TELEBRAS in Campinas, developing methods for video quality assessment. His experience as professor of Computer Science and Engineering started in 1998 at the University of Campinas. His main research interests include image/video processing, 
visualization, and analysis; graph algorithms and dynamic programming; image annotation, organization, and retrieval; machine learning and pattern recognition; and image analysis applications in Biology, Medicine, Biometrics, Geology, and Agriculture.
\end{IEEEbiographynophoto}


\vfill


\end{document}